\documentclass[11pt]{article}

\usepackage[final]{acl}

\usepackage{microtype}
\usepackage{graphicx}
\usepackage{booktabs}
\usepackage{amsmath}
\usepackage{amssymb}

\usepackage{colortbl}
\usepackage{xcolor}
\usepackage{multirow}

\usepackage{booktabs}
\usepackage{array}
\usepackage{ragged2e}

\usepackage[english,bidi=default]{babel} % English as the main language.
\babelfont{rm}{TeXGyreTermesX} % similar to Times
\babelprovide[import]{hindi}
\babelfont[*devanagari]{rm}{Lohit Devanagari}
\babelprovide[import]{arabic}
\babelfont[*arabic]{rm}{Noto Sans Arabic}

\usepackage{tcolorbox}
\usepackage{adjustbox}
\usepackage{fvextra}

\usepackage{enumitem}
\usepackage{pgfplots}
\pgfplotsset{compat=1.17}
\tcbuselibrary{breakable}

\DefineVerbatimEnvironment{PromptCode}{Verbatim}{%
  breaklines=true,
  breakanywhere=true,
  breaksymbolleft={},
  breaksymbolright={},
  fontsize=\footnotesize
}

\definecolor{nmgray}{RGB}{229,229,229}

\tcbset{
    colback=gray!5,
    colframe=black!75!black,
    before skip=2pt, % Adjust this value as needed
    after skip=2pt, % Adjust this value as needed
    left=0pt,
    right=0pt,
    top=0pt,
    bottom=0pt,
    boxsep=5pt, % Adjust this value to change the space between the frame and the content inside the box
    arc=0pt, % Adjust if you want to have rounded corners
    boxrule=1pt, % Adjust the thickness of the frame
    before upper={
    \setlength{\baselineskip}{\normalbaselineskip}
    \setlength{\parskip}{0.5\baselineskip}
  }
}

\newtcolorbox{mybox}[2][]{
width=\columnwidth,
colback = nmgray!75!white, 
colframe = nmgray!75!white, 
boxsep=0pt,left=9pt,right=10pt,top=0pt,bottom=0pt,
fontupper=\linespread{0.9}\selectfont,
title=#2,#1}

\title{PaperScout: An Autonomous Agent for Academic Paper Search with Process-Aware Sequence-Level Policy Optimization}

\author{%
  Tingyue Pan, Jie Ouyang, Mingyue Cheng\thanks{Corresponding author.}, Qingchuan Li, \\\textbf{ Zirui Liu, Daoyu Wang, Mingfan Pan, Shuo Yu, Qi Liu, Enhong Chen}
   \\[0.3em]
   State Key Lab of Cognitive Intelligence, University of Science and Technology of China \\ 
  \texttt{\{pty12345,ouyang\_jie,chouli,liuzirui,wdy030428,mfpan,yu12345\}@mail.ustc.edu.cn} \\
  \texttt{\{mycheng,qiliuql,cheneh\}@ustc.edu.cn} \\
  }

\begin{document}

\maketitle
\begin{abstract}

Academic paper search is a fundamental task in scientific research, yet most existing approaches organize retrieval around predefined workflows or structured interaction protocols that struggle with complex, conditional queries. To address this limitation, we propose PaperScout, an autonomous agent that reformulates paper search as a sequential decision-making process. Unlike static workflows, PaperScout dynamically decides whether, when, and how to invoke \texttt{search} and \texttt{expand} tools based on accumulated retrieval context. However, training such agents presents a fundamental challenge: standard reinforcement learning methods, typically designed for single-turn tasks, suffer from a granularity mismatch when applied to multi-turn agentic tasks—where token-level optimization diverges from the granularity of sequence-level interactions—leading to noisy credit assignment and unstable training dynamics. We introduce Proximal Sequence Policy Optimization (PSPO), a process-aware, sequence-level policy optimization method that aligns optimization with agent--environment interaction. Comprehensive experiments on both synthetic and real-world benchmarks demonstrate that PaperScout significantly outperforms strong structured retrieval and RL baselines in both recall and relevance, validating the effectiveness of our adaptive agentic framework and optimization strategy. Our code is available\footnote{https://github.com/AgentR1/PaperScout}.

\end{abstract}

% \vspace{-0.1in}

\section{Introduction}

% \vspace{-0.09in}

Academic paper search underpins effective knowledge discovery \cite{timmins2005conduct, marchionini2006exploratory}. Most traditional approaches rely on lexical or semantic matching over largely static corpora, which perform well on well-formed queries \cite{vine2006google, zhu2025large, shi2025spar}. However, as scholarly publications originate from increasingly heterogeneous sources, static databases become costly to maintain \cite{manghi2024challenges}. Meanwhile, traditional methods exhibit limited capability when faced with fine-grained and conditional queries \cite{gusenbauer2020academic}. A researcher seeking \textit{``Apply reinforcement learning to protein folding while excluding Transformer-based architectures"} often encounters thematically relevant but constraint-violating papers, leading to substantial manual search costs \cite{gusenbauer2021every}.

% \vspace{0.1in}

\begin{figure}[t]
  \centering
  \includegraphics[width=\linewidth]{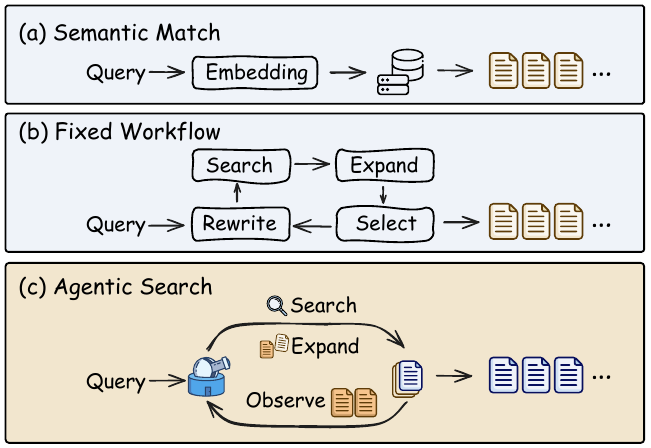}
  \caption{Comparison of different academic search paradigms: Semantic Match, Fixed Workflow, and Agentic Search.}
  \label{fig:motivation}
\end{figure}

% \vspace{-0.02in}

% Recent advances in large language models (LLMs) have introduced new opportunities for academic paper search \cite{zhang2025scientific,wang2025paperarena}. Going beyond single-round querying and static ranking, LLMs leverage accumulated context to guide subsequent steps and produce structured outputs, such as rewritten queries, candidate inspection criteria, and citation cues \cite{zhu2025large}, thereby enabling fine-grained control over the retrieval process. Unlike traditional semantic matching, this interaction-driven retrieval better reflects practical search processes, effectively handling complex queries and gradually narrowing the search space \cite{shi2025spar}.

Recent advances in large language models (LLMs) have introduced new opportunities for academic paper search \cite{zhang2025scientific,wang2025paperarena}. Beyond single-round querying and static ranking, LLMs leverage accumulated context to guide subsequent decisions \cite{cheng2026comprehensive} and produce structured outputs, such as rewritten queries, candidate inspection criteria, and citation cues \cite{zhu2025large}, enabling fine-grained control over retrieval. Unlike traditional semantic matching, this interaction-driven retrieval better reflects search processes, handling complex queries and narrowing the search space \cite{shi2025spar}.

% Beyond single-round querying and static ranking, LLMs leverage accumulated context to guide subsequent steps and produce structured outputs, such as rewritten queries, candidate inspection criteria, and citation cues \cite{cheng2026comprehensive}, enabling fine-grained control over retrieval.

Despite the increased expressive capacity of LLM-based approaches for academic paper search, most systems still rely on predefined workflows or constrained interaction protocols. They decompose retrieval into stages such as query rewriting, search, paper inspection, and reference expansion, and specify how these stages are connected. This constraint also applies to agentic systems such as PaSa, where search, expansion, and stopping actions are still organized by a paper-queue traversal protocol. Consequently, learning mainly improves local decisions within a fixed structure, but leaves higher-level search control—when to broaden search, deepen expansion, or redirect exploration—largely underexplored.

% Despite the increased expressive capacity of LLM-based approaches for academic paper search, most systems still organize retrieval around predefined workflows or constrained interaction protocols. These approaches decompose retrieval into stages such as query rewriting, search, paper inspection, and reference expansion, and specify how these stages are connected. Even agentic systems such as PaSa support search, expansion, and stopping actions, their behavior is still shaped by a paper-queue traversal protocol. As a result, learning mainly optimizes local actions within a fixed structure, leaving broader control decisions—such as when to broaden search, deepen expansion, or redirect exploration—underexplored.

% Even agentic systems such as PaSa can invoke search, expansion, and stopping actions, but their high-level behavior is still shaped by a paper-queue traversal protocol that populates candidate papers and then explores citation neighborhoods. As a result, learning mainly optimizes local actions within a prescribed structure, while the broader control problem of when to broaden search, deepen expansion, or shift exploration direction remains underexplored.

To address these limitations, we propose \textbf{PaperScout}, an autonomous paper search agent that reframes retrieval as a sequential decision-making process rather than predefined workflow execution. Instead of being tied to a search-then-traverse protocol, PaperScout decides at each step \emph{whether}, \emph{when}, and \emph{how} to invoke \texttt{Search} and \texttt{Expand} based on the papers accumulated so far. By giving the agent direct control over retrieval decisions,
PaperScout enables flexible, context-dependent exploration, allowing search strategies to evolve as the retrieval process unfolds.

% , enabling flexible, context-dependent exploration as retrieval unfolds.

% However, training such an autonomous multi-turn retrieval agent presents a dilemma for existing policy optimization paradigms. On one hand, outcome-oriented methods (e.g., GRPO) \cite{shao2024deepseekmath}, which have proven effective for reasoning tasks, typically treat the entire interaction trajectory as a single unit. The granularity is coarse for multi-turn retrieval, where intermediate feedback is essential to guide the agent through sequential decisions. On the other hand, standard token-level approaches (e.g., PPO) \cite{schulman2017proximal} operate at a granularity that is misaligned with the agent's turn-based interactions, often leading to unstable value estimation and noisy credit assignment.

However, training such an autonomous multi-turn retrieval agent remains challenging for existing policy optimization methods. Outcome-oriented approaches (e.g., GRPO) \cite{shao2024deepseekmath}, while effective for reasoning tasks, primarily optimize against final outcomes and cannot naturally incorporate process rewards from intermediate retrieval steps. This limitation is particularly restrictive for multi-turn retrieval, where stepwise feedback is crucial for guiding sequential decisions. In contrast, standard token-level methods (e.g., PPO) \cite{schulman2017proximal} optimize policies at the token level, whereas the agent’s decisions are made at the interaction level, with each turn producing a complete response that triggers tool executions. This mismatch leads to misaligned optimization, unstable value estimation, and noisy credit assignment.

To bridge this granularity mismatch, we propose \textbf{Proximal Sequence Policy Optimization} (PSPO), a process-aware policy optimization method tailored for multi-turn agents. Unlike outcome-only baselines, PSPO performs advantage estimation at the sequence level (per interaction turn) and explicitly incorporates intermediate process rewards, ensuring that the optimization granularity aligns with agent–environment interactions.

% To bridge this granularity mismatch, we propose \textbf{Proximal Sequence Policy Optimization} (PSPO), a process-aware policy optimization method tailored for multi-turn agents. Unlike outcome-only baselines, PSPO performs advantage estimation at the sequence level (per interaction turn) and explicitly incorporates process feedback, ensuring that the optimization granularity aligns strictly with the agent–environment interaction.

In summary, our main contributions are:

\vspace{-0.05in}

\begin{itemize}\setlength\itemsep{0.1pt}
    \item We propose PaperScout, the first autonomous paper search agent that adaptively controls search and reference expansion based on accumulated retrieval context.
    \item We introduce Proximal Sequence Policy Optimization (PSPO), a process-aware, sequence-level policy optimization method that aligns optimization granularity with multi-turn agent interactions.
    \item Extensive experiments demonstrate that PaperScout enables more effective and flexible multi-turn retrieval than baselines, while analysis shows that PSPO delivers superior sample efficiency and stability over token-level and outcome-oriented methods.
\end{itemize}

\section{Related Work}

\subsection{Query-Centric Academic Search}
Traditional academic paper search methods predominantly follow a query-centric retrieval setting, in which retrieval is driven by a single user query and focuses on single-round query–document relevance modeling \cite{schutze2008introduction,guo2020deep}. Early approaches retrieve papers by measuring term-level similarity between the query and documents \cite{qaiser2018text}. To mitigate the limitations of semantic mismatch, later work moves toward dense retrieval, which improves query–document matching by encoding queries and documents into semantic vector representations \cite{jiang2019semantic, reimers2019sentence}. More recently, LLMs have been employed to rewrite or expand queries prior to retrieval in order to better capture user intent \cite{ma2023query, gusenbauer2020academic}. Despite these advances, such methods still treat paper search as a single-round query–document matching task, which limits their ability to handle complex search queries \cite{gusenbauer2020academic}.

\subsection{Multi-Turn Academic Paper Search}
To move beyond the query-centric, single-round retrieval paradigm, recent studies have explored multi-turn approaches for academic paper search \cite{cheng2025openlens,aytar2025synergistic,park2025chain}. These methods typically formulate retrieval as a sequence of operations, enabling systems to iteratively search, inspect candidate papers, and expand along citation links. Representative work such as PaSa \cite{he2025pasa} trains a crawler to invoke search, expansion, and stop actions within a paper-queue interaction framework, while SPAR \cite{shi2025spar} integrates query rewriting, reference exploration, and re-ranking in a modular workflow. Despite their effectiveness, these approaches still rely on predefined interaction structures, offering limited flexibility in adapting high-level retrieval strategies to different queries or evolving search contexts.

\subsection{Reinforcement Learning for LLM Agents}

Reinforcement learning has been widely adopted to enhance LLM capabilities \cite{zhou2025time,wang2026steppo, liu2026towards}. Traditional approaches like PPO \cite{schulman2017proximal} operate at the token level, where the difficulty of accurate credit assignment often leads to optimization instability under sparse rewards \cite{ouyang2025token}. To mitigate this, outcome-oriented methods such as GRPO \cite{shao2024deepseekmath} and GSPO \cite{zheng2025group} have gained traction by leveraging group-wise outcome comparisons. However, extending these paradigms to multi-turn agents presents significant challenges \cite{ragen,cheng2025agent}. Most existing approaches generalize outcome-based logic to the trajectory level, treating the entire trajectory as a monolithic unit \cite{luo2025agent}. This granularity is coarse for multi-turn retrieval, as it overlooks the dense process signals available at intermediate steps.

\section{Methodology}

\begin{figure*}[t]
  \centering
  \includegraphics[width=0.98\linewidth]{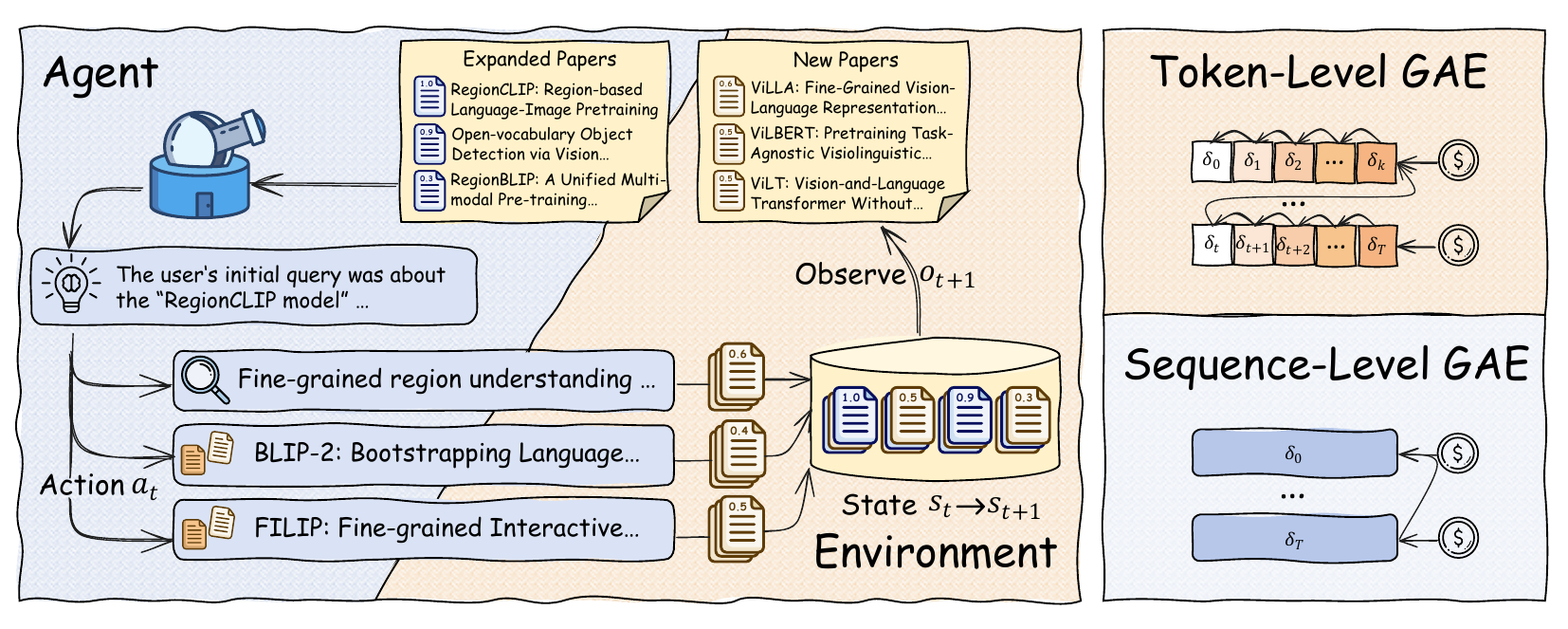}
  \caption{\textbf{Overview of PaperScout and the motivation for PSPO. Left:} PaperScout models multi-turn paper search as a POMDP by maintaining a paper pool as the latent state; at each step, the agent observes a summarized view of the pool and issues \texttt{Search} or \texttt{Expand} tool calls to retrieve new papers, which are then merged to update the pool. \textbf{Right:}  \textit{Granularity mismatch} in credit assignment: PPO assigns single step reward to many tokens, diluting supervision, whereas PSPO treats each complete response as an atomic action and optimizes at the sequence level.}
  \label{fig:main}
\end{figure*}

In this section, we present PaperScout, an autonomous LLM-based agent for academic paper search.
An overview of the PaperScout framework is shown in Figure~\ref{fig:main}.
We first formalize paper search as a Partially Observable Markov Decision Process (POMDP) in Section~\ref{sec:problem_def}.
Building on this formulation, Section~\ref{sec:paperscout_design} instantiates each POMDP component in a concrete multi-tool agent system.
To optimize the agent policy under sequence-level feedback, Section~\ref{sec:pspo} introduces Proximal Sequence Policy Optimization (PSPO).

\subsection{Problem Definition}
\label{sec:problem_def}

We formalize academic paper search as a Partially Observable Markov Decision Process (POMDP)
$\langle \mathcal{S}, \mathcal{A}, \mathcal{P}, \mathcal{R}, \Omega, \mathcal{O}, \gamma \rangle$.
The state $s_t \in \mathcal{S}$ represents a latent \textit{paper pool} that contains all papers accumulated up to step $t$.
Due to limited context, the agent cannot access $s_t$ directly and instead receives an observation
$o_t=\mathcal{O}(s_t)\in\Omega$, which provides a partial view of the pool (e.g., a small subset of highly relevant papers).
Based on $o_t$, the agent chooses an action $a_t \in \mathcal{A}$ to acquire new information.
In our setting, an action is instantiated by invoking one or more external retrieval tools (such as web search or reference expansion), whose execution returns additional candidate papers.
The transition $\mathcal{P}(s_{t+1}\mid s_t,a_t)$ updates the paper pool by incorporating newly retrieved papers, and the reward
$r_t=\mathcal{R}(s_t,a_t)$ measures the marginal utility of this acquisition with respect to the user query.
This formulation casts paper search as an iterative decision-making problem rather than a one-shot ranking task.

\subsection{PaperScout}
\label{sec:paperscout_design}

PaperScout instantiates the POMDP in a modular agentic system (Figure~\ref{fig:main}).
While Section~\ref{sec:problem_def} provides an abstract formulation, we now describe the concrete design for state tracking, observation construction, and policy execution.

\begin{table}[t]
    \centering
    \small
    \begin{tabular}{lp{0.6\linewidth}}
        \toprule
        \textbf{Tool} & \textbf{Description} \\
        \midrule
        \texttt{Search(query)} & Calls scholarly search APIs to retrieve new papers. \\
        \texttt{Expand(paper)} & Retrieves reference papers cited by the input paper. \\
        \bottomrule
    \end{tabular}

    \caption{Tools available to the agent.}
    \vspace{-0.1in}
    \label{tab:tool-functions}
\end{table}

\paragraph{State and Observation Space.}
We implement the latent state $s_t$ as a \textit{paper pool} $\mathcal{B}_t$.
Each paper in $\mathcal{B}_t$ stores (i) its content and metadata, (ii) a relevance score $\rho(\cdot)$ with respect to the user query, and (iii) an expansion flag indicating whether its references have been explored.
Together, $\mathcal{B}_t$ maintains the agent's current search frontier.

The observation $o_t$ summarizes $\mathcal{B}_t$ under the agent's limited context budget.
We use a dual-list view that partitions top-ranked papers into (a) expanded papers, which provide stable context from previously explored nodes, and (b) unexpanded papers, which are candidates for further exploration.
We further augment $o_t$ with the interaction history $\mathcal{H}_{t-1}$ to reduce redundant exploration, including past queries and previously expanded papers.

\paragraph{Policy and Execution Mechanism.}
The environment provides a structured observation $o_t$, which we serialize into the LLM input sequence $x_t$ using a fixed prompt template together with the user query and tool description.
The LLM then samples an output sequence $y_t \sim \pi_\theta(\cdot \mid x_t)$.
We parse $y_t$ into $y_t=(z_t,\mathcal{C}_t)$, where $z_t$ is a brief reasoning trace and $\mathcal{C}_t$ is a set of tool invocations defined in Table~\ref{tab:tool-functions}.
The environment executes $\mathcal{C}_t$ and returns raw results $\mathcal{D}(\mathcal{C}_t)$.
We then filter these results to obtain the accepted candidate set $\mathcal{V}_t$:
\begin{equation}
    \mathcal{V}_{t} = \{\, p \in \mathcal{D}(\mathcal{C}_t)\ \mid\ \rho(p) \ge \tau\ \land\ p \notin \mathcal{B}_{t}\},
\end{equation}
where $\tau$ is a minimum relevance threshold.
Only papers in $\mathcal{V}_t$ are merged into the pool, yielding the state transition $s_t \to s_{t+1}$.

% \vspace{-0.1in}

\begin{figure*}[t]
  \centering
  \includegraphics[width=\linewidth]{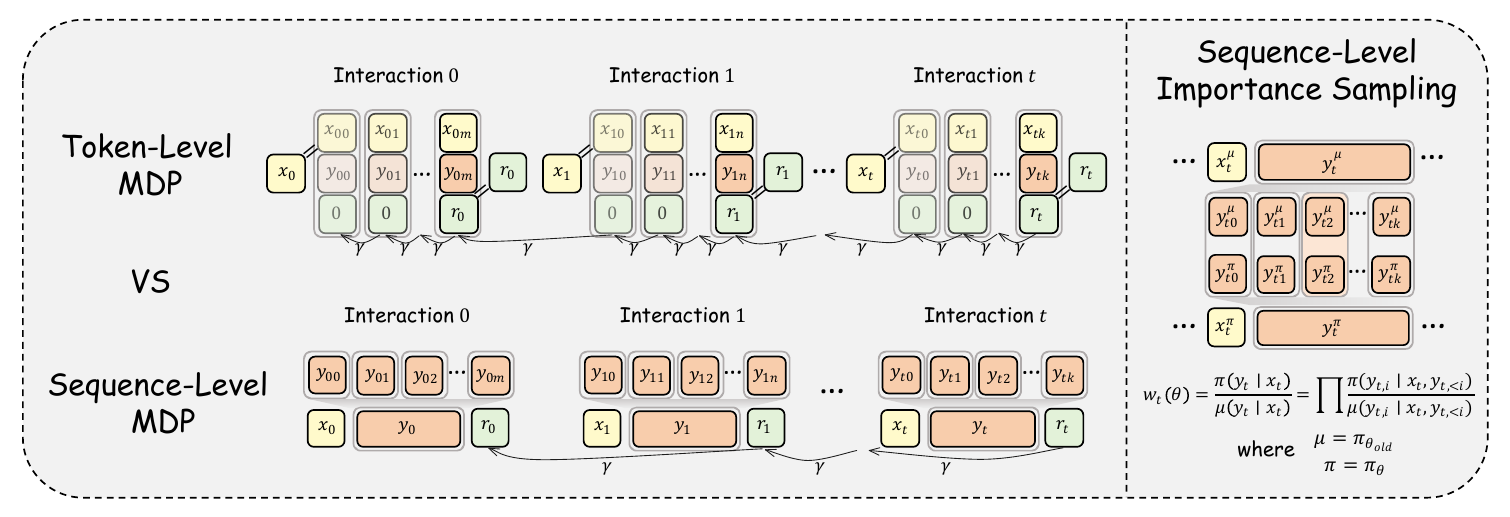}
  \caption{\textbf{PSPO optimizes policies at the sequence (interaction) level.}
  PPO assigns a single turn-level reward to many tokens in the generated response, which dilutes learning signals under token-level optimization. PSPO instead treats each complete response $y_t$ as the atomic action in an interaction-level MDP, and performs advantage estimation and policy updates at the sequence level. This interaction-level formulation naturally leads to sequence-level importance sampling for off-policy correction.}
  \label{fig:pspo}
\end{figure*}

\subsection{Proximal Sequence Policy Optimization}
\label{sec:pspo}

To effectively train the agent within the POMDP framework, we introduce Proximal Sequence Policy Optimization (PSPO). This section details the reward formulation and the core optimization algorithm designed to bridge the gap between token-level generation and sequence-level utility.

% \vspace{-0.1in}

\paragraph{Reward Formulation.}
PaperScout aims to maximize the recall of relevant papers.
We design the reward so that the expected return matches the expected number of relevant papers acquired over the search process.
Let $\mathcal{V}_t$ denote the set of novel papers accepted into the pool at step $t$.
To reduce variance, we compute the gain on the top-$k$ papers in $\mathcal{V}_t$ ranked by relevance (using all papers if $|\mathcal{V}_t|<k$).
The step reward is:
\begin{equation}
    r_t =
    \underbrace{\sum_{p \in \text{top-}k(\mathcal{V}_t)} \rho(p)}_{\text{relevance gain}}
    \;-\;
    \underbrace{\eta \sum_{c \in \mathcal{C}_t} \mathbb{I}(c \in \mathcal{H}_{t-1})}_{\text{repetition penalty}},
\end{equation}
where $\rho(p)\in[0,1]$ is the relevance score of paper $p$, interpreted as the probability that $p$ is relevant to the user query.
With this interpretation, the cumulative relevance gain corresponds to the expected number of relevant papers retrieved, while the repetition term discourages redundant tool usage.

\paragraph{Algorithm Implementation.}
In our setting, the agent interacts with the environment at the \emph{sequence level}: each step corresponds to generating a complete response that triggers tool executions and yields a single step reward.
This induces a granularity mismatch for token-level optimization, as illustrated in Figure~\ref{fig:pspo}: sparse sequence-level feedback must be attributed to tokens, leading to noisy credit assignment and unstable value learning.

PSPO resolves this mismatch by optimizing policies at the sequence level.
Let $x_t$ be the prompt obtained by serializing the observation $o_t$, and $y_t$ be the complete response at step $t$.
For a trajectory $\mathcal{T}=\langle x_0,y_0,\dots,x_{T-1},y_{T-1}\rangle$ of length $T$, we define $r_t=r(x_t,y_t)$ and learn a value function $V_\phi(x_t)$.
We estimate advantages with GAE:

\begin{equation} 
\hat{A}_t = \sum_{l=0}^{T-t-1} (\gamma\lambda)^l\, \delta_{t+l},
\end{equation}

\noindent where $\delta_t = r(x_t,y_t) + \gamma V_\phi(x_{t+1}) - V_\phi(x_t) $ denotes the temporal-difference (TD) error.
% with $V_\phi(x_T)=0$.

The actor is trained with a clipped surrogate objective function:
\begin{equation}
    \begin{aligned}
        \mathcal{L}_{\text{actor}}&(\theta) =
        \mathbb{E}_{\mathcal{T}\sim \pi_{\theta_{\text{old}}}}
        \Big[\min\big(w_t(\theta)\hat A_t, \\
        & \text{clip}(w_t(\theta), 1-\varepsilon_{low},1+\varepsilon_{high})\hat A_t\big)\Big],
    \end{aligned}
\end{equation}
where the sequence-level importance ratio is
\begin{equation}
\begin{aligned}
w_t(\theta)
&= \frac{\pi_\theta(y_t\mid x_t)}{\pi_{\theta_{\text{old}}}(y_t\mid x_t)} \\
&= \prod_{i=1}^{L_t}
\frac{\pi_\theta(y_{t,i}\mid x_t, y_{t,<i})}
{\pi_{\theta_{\text{old}}}(y_{t,i}\mid x_t, y_{t,<i})}.
\end{aligned}
\end{equation}
The critic minimizes $\mathbb{E}\big[(V_\phi(x_t)-R_t)^2\big]$ with
\begin{equation}
R_t = \sum_{k=0}^{T-t-1}\gamma^k r_{t+k}.
\end{equation}

\section{Experiments}

\begin{table*}[t]
\centering
\caption{Performance comparison on the RealScholarQuery and AutoScholarQuery benchmarks. Best results are highlighted in \textbf{bold} and second-best results are \underline{underlined}. Note that PaperScout develops based on Qwen3-4B.}
\label{tab:main_tab}

\resizebox{\textwidth}{!}{
\begin{tabular}{l|ccc|cccc|c}
\toprule
\multirow{2}{*}{\textbf{Model}}
& \multicolumn{3}{c|}{\textbf{Post-threshold}}
& \multicolumn{4}{c|}{\textbf{Recall@k}}
& \multirow{2}{*}{\textbf{LLM-score}} \\
\cmidrule{2-8}
& Precision & F1 & Recall 
& Recall@all & Recall@100 & Recall@50 & Recall@25 & \\
\midrule

\multicolumn{9}{c}{\textbf{RealScholarQuery}} \\
\midrule
Google Search & 0.059 & 0.074 & 0.304 & 0.304 & 0.304 & 0.267 & 0.221 & 1.116 \\
Google Search + LLM & 0.067 & 0.077 & 0.254 & 0.254 & 0.254 & 0.216 & 0.202 & 1.237 \\
Google Scholar & 0.045 & 0.064 & 0.247 & 0.247 & 0.247 & 0.205 & 0.158 & 1.251 \\
PaSa & 0.415 & 0.417 & 0.541 & 0.598 & 0.591 & 0.572 & 0.517 & 2.111 \\
SPAR & 0.412 & 0.408 & 0.496 & 0.545 & 0.545 & 0.537 & 0.504 & 2.415 \\
\midrule
PaperScout-Qwen3-4B & 0.404 & 0.411 & 0.497 & 0.545 & 0.543 & 0.535 & 0.498 & 2.261 \\
PaperScout-Qwen3-Max & \underline{0.435} & \underline{0.427} & \underline{0.562} & \underline{0.664} & \underline{0.648} & \underline{0.606} & \underline{0.532} & \underline{2.483} \\
\rowcolor{gray!15}
PaperScout
 & \textbf{0.442} & \textbf{0.441} & \textbf{0.574}
 & \textbf{0.691} & \textbf{0.683} & \textbf{0.646} & \textbf{0.577}
 & \textbf{2.576} \\
\midrule

\multicolumn{9}{c}{\textbf{AutoScholarQuery}} \\
\midrule
Google Search
 & 0.009 & 0.017 & 0.127 & 0.127 & 0.127 & 0.101 & 0.058 & 1.263 \\
Google Search + LLM
 & 0.010 & 0.019 & 0.138 & 0.138 & 0.138 & 0.115 & 0.081 & 1.322 \\
Google Scholar
 & 0.006 & 0.011 & 0.081 & 0.081 & 0.081 & 0.070 & 0.057 & 1.212 \\
PaSa
 & 0.095 & 0.129 & \underline{0.442} & 0.699 & 0.579 & 0.463 & \underline{0.344} & 2.186 \\
SPAR
 & 0.091 & \underline{0.131} & 0.386 & 0.561 & 0.535 & 0.466 & 0.341 & 2.295 \\
\midrule
PaperScout-Qwen3-4B
 & 0.092 & 0.128 & 0.382 & 0.564 & 0.540 & 0.448 & 0.324 & 2.202 \\
PaperScout-Qwen3-Max
 & \underline{0.102} & 0.115 & 0.427 & \underline{0.770} & \underline{0.584} & \underline{0.469} & 0.336 & \underline{2.368} \\
\rowcolor{gray!15}
PaperScout
 & \textbf{0.115} & \textbf{0.134} & \textbf{0.459}
 & \textbf{0.811} & \textbf{0.636} & \textbf{0.496} & \textbf{0.357}
 & \textbf{2.467} \\
\bottomrule
\end{tabular}
}
\end{table*}

\subsection{Experiment Setup}

We provide a clear basis for our experiments, including the datasets, baselines, key experimental environments, and evaluation metrics.

\paragraph{Dataset Description.} We conduct experiments on two benchmarks: AutoScholarQuery and RealScholarQuery \cite{he2025pasa}, which represent synthetic and expert-curated scholarly search scenarios. AutoScholarQuery is a large-scale synthetic benchmark constructed from recent top-tier conference papers, including ICLR, ICML, NeurIPS, ACL, and CVPR, while RealScholarQuery is a human-annotated benchmark designed to evaluate realistic and challenging scholarly search settings. Detailed statistics of the datasets are summarized in Table~\ref{tab:datasets}.

\begin{table}[ht]
    \centering
    \resizebox{0.95\linewidth}{!}{
    \begin{tabular}{p{0.45\linewidth}ccc}
        \toprule
        \textbf{Dataset} & \textbf{\#Train} & \textbf{\#Dev} & \textbf{\#Test} \\
        \midrule
        AutoScholarQuery & 33551 & 1000 & 1000 \\
        RealScholarQuery & -- & -- & 50 \\
        \bottomrule
    \end{tabular}
    }
    \caption{Details of the two datasets.}
    \label{tab:datasets}
\end{table}

For reinforcement learning, the agent is trained on the AutoScholarQuery training split and validated on the development split.
We evaluate on two test sets: (i) a filtered subset of the AutoScholarQuery test set with 112 queries having at least five ground-truth relevant papers, and (ii) RealScholarQuery, which contains 50 real-world scholarly queries with reference papers.

\paragraph{Retrieval Backends.}
To enable stable RL training under high concurrency, we build a fully local retrieval environment for training.
Specifically, we deploy Milvus\footnote{\url{https://milvus.io}} over millions of paper metadata to emulate a scholarly search service, and pre-cache millions of papers from ar5iv\footnote{\url{https://ar5iv.labs.arxiv.org}} to support reference-based expansion.
If a requested paper is missing from the local cache, it would be downloaded on demand from ar5iv or arXiv\footnote{\url{https://arxiv.org}}.
During evaluation, we standardize the search backend by issuing search calls through Google Search for fair comparison across methods.

\vspace{-0.03in}

\paragraph{Baselines.}
We compare against a diverse set of baselines spanning traditional academic search engines, LLM-enhanced retrieval workflows, and recent multi-turn paper search systems:

\vspace{-0.03in}

\begin{itemize}
    \setlength{\itemsep}{0.5pt}
    \item \textbf{Google Search \footnote{\url{https://serper.dev}}.} Standard Google Search using the original query.
    \item \textbf{Google Search + LLM.} Google Search with a refined query generated by LLM.
    \item \textbf{Google Scholar.} Direct retrieval from Google Scholar without LLM intervention.
    \item \textbf{PaSa \cite{he2025pasa}.} A reinforcement
learning–based paper search system that optimizes query rewriting and reference expansion under a predefined workflow.
    \item \textbf{SPAR \cite{shi2025spar}.} A multi-turn academic search framework that integrates multiple retrieval components within a predefined structured workflow.
\end{itemize}

\paragraph{Our Variants.}
We evaluate three variants of our approach to disentangle the effects of RL fine-tuning and model scale:
(i) PaperScout (Qwen3-4B fine-tuned with PSPO),
(ii) PaperScout-Qwen3-4B (the same backbone without RL fine-tuning),
and (iii) PaperScout-Qwen3-Max (a larger backbone, Qwen3-Max, without RL fine-tuning).

\vspace{-0.02in}

\paragraph{Evaluation Protocol.}
For Google Search, Google Search + LLM, and Google Scholar, we evaluate the top 100 retrieved papers.
For other multi-turn methods, we use each system's reranked and filtered outputs.
For PaperScout, candidate papers are ranked by relevance score $\rho(\cdot)$, and those with score $\rho(p)\ge 0.5$ are retained.

\vspace{-0.02in}

\paragraph{Metrics.}
% We report Precision, Recall, F1-score, and an LLM-based relevance score.
% Precision, Recall, and F1-score are computed by exact matches between retrieved papers and ground-truth relevant papers.
We report Precision, Recall, F1-score, Recall@k, and an LLM-based relevance score. The post-threshold Precision, Recall, and F1-score are computed based on exact matches between the ground-truth relevant papers and the set of papers retained after applying a relevance threshold $\rho(p)\ge 0.5$. Recall@k measures the fraction of ground-truth relevant papers appearing within the top-k ranked results.

To mitigate evaluation bias caused by incomplete annotations, we additionally report an LLM-based relevance score. Three LLMs (DeepSeek-V3.2 \cite{liu2025deepseek}, Qwen3-Max \cite{yang2025qwen3}, and GPT-5.1) independently rate each retrieved paper–query pair on a four-level scale (0–3), where higher scores indicate greater relevance. The final score is the average of the three scores. Details of the scoring protocol and reliability analysis are provided in Appendix~\ref{sec:llm_score}.

% To mitigate evaluation bias caused by incomplete annotations, we additionally report an LLM-based score: three LLMs (DeepSeek-V3.2 \cite{liu2025deepseek}, Qwen3-Max \cite{yang2025qwen3}, and GPT-5.1) independently judge the relevance between each retrieved paper and the user query.
% Relevance is rated on a four-level scale (0--3), and we average the three ratings to obtain a final score in $[0,3]$.
% Details of the scoring protocol and reliability analysis are provided in Appendix~\ref{sec:llm_score}.

\begin{figure}[t]
  \centering
  \includegraphics[width=\linewidth]{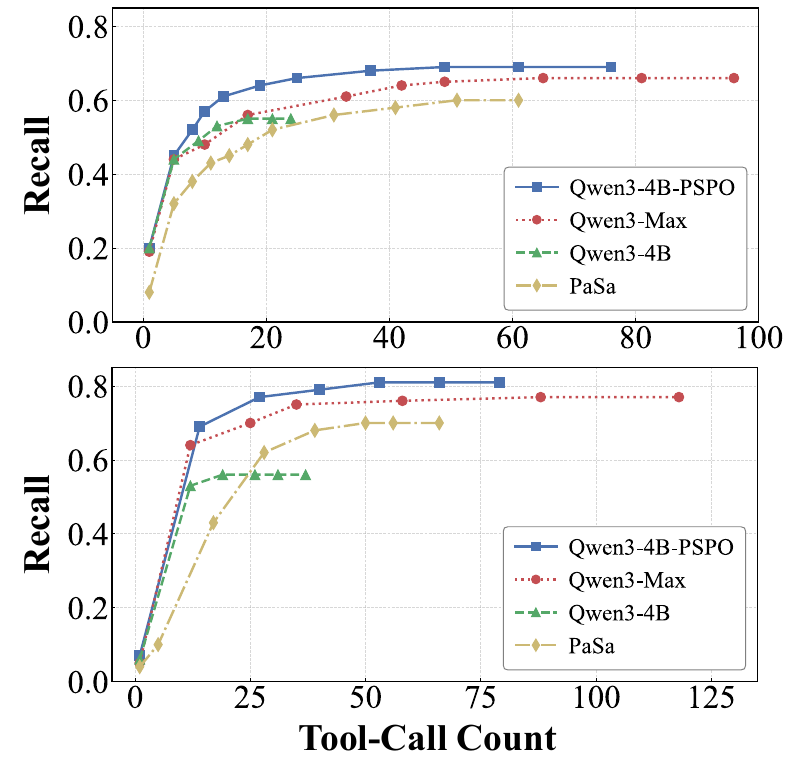}
  \caption{Average recall as tool calls accumulate on RealScholarQuery (top) and AutoScholarQuery (bottom). PSPO consistently yields higher recall with the same number of tool calls.}

  \label{fig:tool_call}
\end{figure}

\vspace{-0.05in}

\subsection{Main Results}
\vspace{-0.05in}
Table~\ref{tab:main_tab} summarizes the performance on RealScholarQuery and AutoScholarQuery.
Across both benchmarks, multi-turn retrieval methods substantially outperform single-shot baselines, highlighting the importance of iterative tool use for academic paper search under complex query settings.

% Among all multi-turn systems, PaperScout consistently achieves the strongest performance.
% On RealScholarQuery, PaperScout attains the highest Recall of 0.574 (vs.\ 0.541 for the best baseline) together with the best LLM-score of 2.576, and improves the best baseline F1-score from 0.417 to 0.441 (\(+5.8\%\) relative).
% On AutoScholarQuery, PaperScout again ranks first, yielding the highest Recall of 0.459 and highest LLM-score of 2.467, while also achieving the best F1-score (0.134).

Among all multi-turn systems, PaperScout consistently achieves the strongest performance. On RealScholarQuery, PaperScout attains the highest Recall of 0.574 (vs.\ 0.541 for the best baseline) together with the best LLM-score of 2.576, and improves the best baseline F1-score from 0.417 to 0.441 (\(+5.8\%\) relative). PaperScout also achieves the best Recall@k across all cutoffs, indicating stronger ranking quality of relevant papers. On AutoScholarQuery, PaperScout again ranks first, yielding the highest Recall of 0.459 and highest LLM-score of 2.467, while also achieving the best F1-score (0.134). Notably, the RL-trained 4B PaperScout matches or surpasses a larger untrained backbone (PaperScout-Qwen3-Max), motivating further analysis of its tool-call efficiency and optimization behavior in the following sections.

\begin{figure*}[t]
  \centering
  \includegraphics[width=\linewidth]{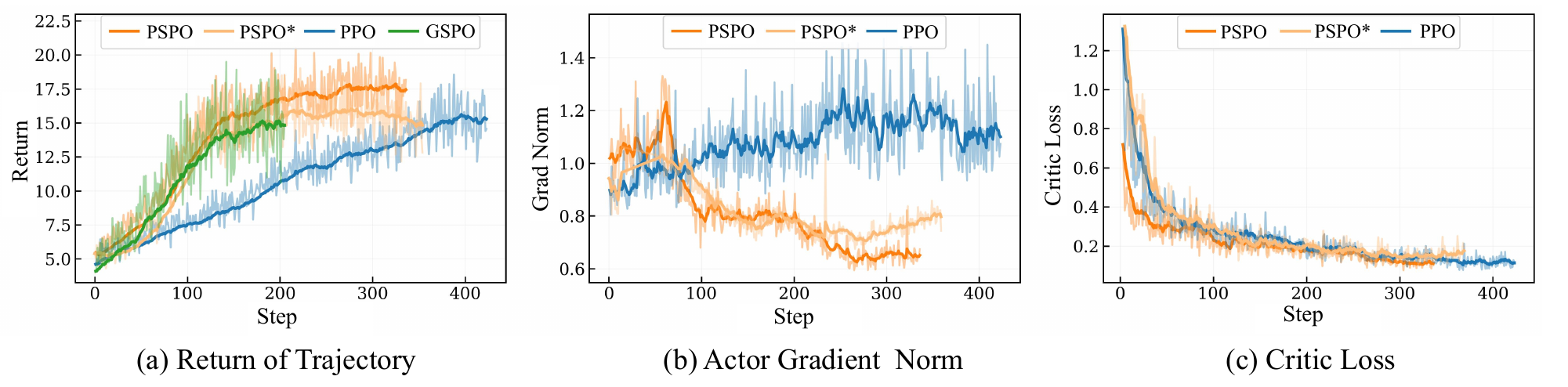}
  \vspace{-0.2in}
  \caption{Training dynamics and optimization statistics for different policy optimization methods. (a) Trajectory returns. PSPO converges faster and reaches a higher plateau than PPO, GSPO, and PSPO$^*$ (PSPO without process rewards). (b) Actor gradient norm during policy updates. PSPO shows smaller gradients with a clearer downward trend than PPO. (c) Critic loss. PSPO is consistently lower than PPO and PSPO$^*$, particularly in early training.}
  \label{fig:rl_curve}
\end{figure*}

\begin{table}[t]
\centering
\vspace{-0.08in}
\caption{Performance comparison of PaperScout optimized by PPO, GSPO and PSPO on RealScholarQuery.}
\vspace{-0.08in}
\label{tab:pspo}
\resizebox{\linewidth}{!}{
\begin{tabular}{l|cccc}
\toprule
\textbf{Model} 
 & Precision & F1 & Recall & LLM-score \\
\midrule
PPO 
 & 0.405 & 0.408 & 0.537 & 2.417\\
GSPO
 & 0.433 & 0.439 & 0.557 & 2.510 \\
\rowcolor{gray!15}
PSPO
 & \textbf{0.442} & \textbf{0.441} & \textbf{0.574} & \textbf{2.576} \\
\bottomrule
\end{tabular}
}
\end{table}

\subsection{Analysis of PaperScout}
\vspace{-0.05in}
To analyze the strong retrieval performance of PaperScout, we examine its behavior from the perspectives of tool-call efficiency and flexibility in multi-turn retrieval.

\paragraph{Tool-Call Efficiency.}
% Figure~\ref{fig:tool_call} illustrates how recall evolves as tool calls accumulate on both benchmarks.
% Overall, PaperScout achieves consistently higher recall than all baselines under the same number of tool calls, indicating a larger retrieval gain per invocation.
% In the low-call regime, the trained PaperScout agent exhibits a steeper increase in recall compared with PaSa and the untrained 4B variant, suggesting more effective early-stage multi-turn retrieval.
% As tool calls increase, the untrained 4B agent quickly saturates, whereas PaperScout continues to accumulate relevant papers and maintains a clear advantage throughout the trajectory.
% Notably, PaperScout matches or surpasses the much larger Qwen3-Max backbone across a wide range of tool-call counts on both datasets, demonstrating that efficient multi-turn retrieval behavior can compensate for model scale through more informative tool usage.

Figure~\ref{fig:tool_call} shows how recall evolves as tool calls accumulate on both benchmarks. PaperScout consistently achieves higher recall than all baselines under the same number of tool calls, indicating greater retrieval gain per invocation. In the low-call regime, the trained PaperScout agent exhibits a steeper recall increase than PaSa and the untrained 4B variant, suggesting more effective early-stage retrieval. As tool calls increase, the untrained 4B agent quickly saturates, whereas PaperScout continues to accumulate relevant papers and maintains a clear advantage. Notably, PaperScout matches or surpasses the much larger Qwen3-Max backbone across a wide range of tool-call counts on both datasets, indicating that efficient multi-turn retrieval can compensate for model scale through more informative tool usage.

\begin{figure}[t]
  \centering
  \includegraphics[width=0.95\linewidth]{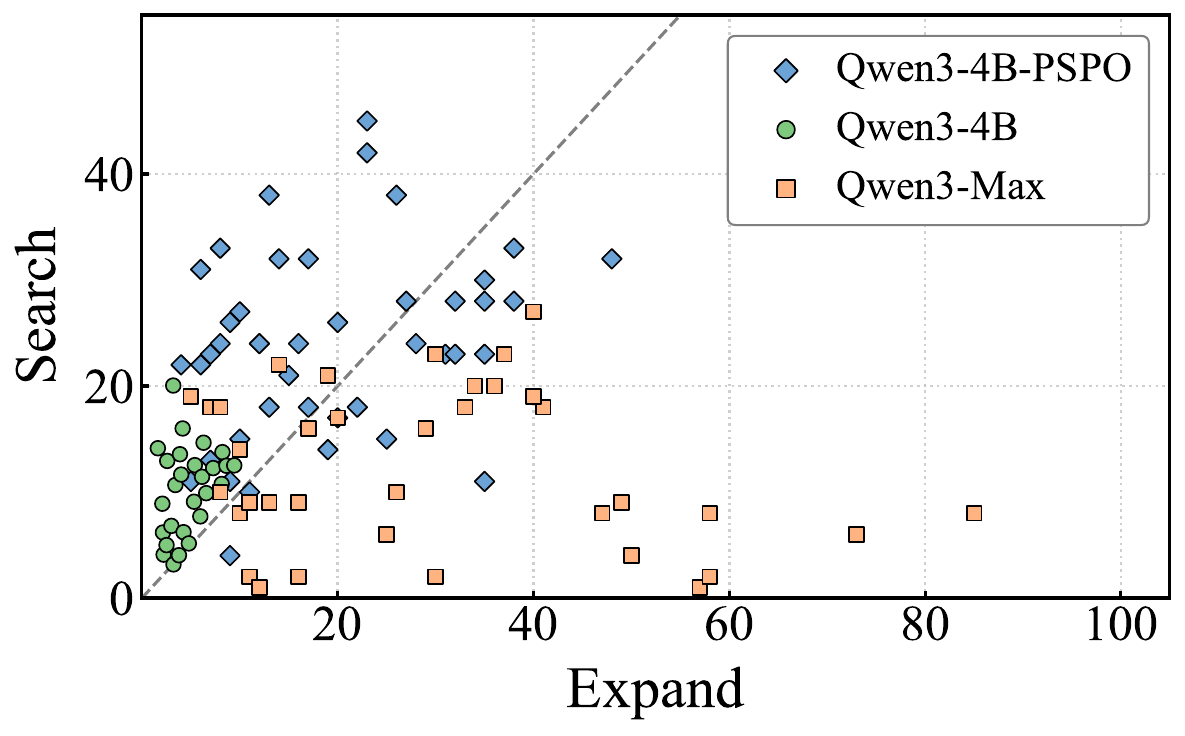}
  \vspace{-0.1in}
  \caption{Tool call distribution across queries for different models.
The trained PaperScout (Qwen3-4B-PSPO) shows a broader, more balanced allocation of \texttt{Search} and \texttt{Expand} calls than the untrained Qwen3-4B and Qwen3-Max, indicating flexible multi-turn retrieval.
}
  \label{fig:tool_distribute}
\end{figure}

\paragraph{Analysis of Tool Distribution.}
Figure~\ref{fig:tool_distribute} analyzes how PaperScout allocates \texttt{Search} and \texttt{Expand} tool calls across queries. The untrained Qwen3-4B agent concentrates near the origin, indicating few tool calls and limited multi-turn exploration. In contrast, Qwen3-Max shows a strong skew toward \texttt{Expand}, with many trajectories dominated by expansion-heavy behavior and relatively few \texttt{Search} calls, restricting new retrieval directions. PaperScout instead occupies a broader region of the \texttt{Search}--\texttt{Expand} space and distributes calls more evenly between the two actions, reflecting a better balance between retrieval breadth and depth. Overall, these patterns suggest that PaperScout enables more flexible multi-turn retrieval rather than fixed or biased tool-use behaviors.

\subsection{Effectiveness of PSPO}
% To evaluate the effectiveness of PSPO as a policy optimization method for multi-turn retrieval, we analyze its impact from two complementary perspectives: final retrieval quality and optimization stability, in comparison with PPO, GSPO and PSPO$^*$ (PSPO without process rewards).

To evaluate PSPO for multi-turn retrieval optimization, we analyze its impact on retrieval quality and optimization stability, compared with PPO, GSPO, and PSPO$^*$ (without process rewards).

\vspace{-0.05in}

\paragraph{Superior Retrieval Performance.}
Table~\ref{tab:pspo} reports the retrieval performance on RealScholarQuery.
PSPO achieves the best results across all metrics, improving Recall to 0.574 (vs.\ 0.557 with GSPO and 0.537 with PPO) and yielding the highest LLM-score of 2.576.
These gains are consistent with the training dynamics shown in Figure~\ref{fig:rl_curve}(a), where PSPO converges faster and reaches a higher return plateau than PPO, GSPO, and PSPO$^*$.

% The slower improvement of PPO stems from the token–sequence granularity mismatch in multi-turn retrieval: token-level optimization under sparse sequence-level feedback amplifies credit assignment noise and makes value fitting more difficult, reducing sample efficiency despite relatively smooth learning curves.
% GSPO instead performs sequence-level optimization and achieves rapid early gains, but exhibits larger return oscillations and an earlier performance ceiling, indicating limited stability when incorporating intermediate process rewards.
% PSPO bridges these behaviors by aligning optimization with sequence-level interaction while leveraging process rewards through the critic, leading to faster convergence and more stable improvements.
% In contrast, PSPO$^*$, which removes process rewards by aggregating them into the final trajectory reward, preserves the same total reward but leads to a lower return plateau, suggesting that step-wise rewards provide useful intermediate learning signals.

% The slower improvement of PPO stems from the token–sequence granularity mismatch in multi-turn retrieval: token-level optimization under sparse sequence-level feedback amplifies credit assignment noise and makes value fitting harder, reducing sample efficiency despite relatively smooth learning curves.

% GSPO performs sequence-level optimization and achieves rapid early gains, but reaches a performance plateau earlier due to its inability to effectively leverage process rewards. Without process-level feedback, it relies on coarse trajectory-level signals, making credit assignment across intermediate decisions difficult and limiting further improvement. 
The slower improvement of PPO stems from the granularity mismatch in multi-turn retrieval: token-level optimization under sparse sequence-level feedback introduces noisy credit assignment and difficult value fitting, reducing sample efficiency despite smooth learning curves.
GSPO performs sequence-level optimization and achieves rapid early gains but plateaus earlier, largely due to its inability to effectively leverage process rewards, relying instead on coarse trajectory-level signals that hinder credit assignment across intermediate decisions.
PSPO bridges these behaviors by aligning optimization with sequence-level interaction while leveraging process rewards through the critic, resulting in faster convergence and more stable improvements. In contrast, PSPO$^*$ aggregates process rewards into the final trajectory reward, preserving the total reward but yielding a lower return plateau, suggesting that step-wise rewards provide useful intermediate learning signals.

\paragraph{Stable Model Optimization.}
Figure~\ref{fig:rl_curve}(b) presents the actor gradient norm during training.
Compared with PPO, PSPO maintains a smaller gradient norm with a clear downward trend, indicating more controlled and stable policy updates under sequence-level optimization.
Figure~\ref{fig:rl_curve}(c) further shows that PSPO consistently achieves a lower critic loss than both PPO and PSPO$^*$, particularly in the early training stage.
This suggests that step-wise process rewards provide more informative learning targets for value regression, whereas removing intermediate rewards weakens credit assignment and makes value estimation more difficult under sparse sequence-level feedback.
Overall, PSPO improves optimization stability for both the actor and critic, leading to stronger and more reliable retrieval performance.

\subsection{Computational Cost and Efficiency}

We analyze the computational cost of training and the runtime efficiency of PaperScout with PSPO.

\paragraph{Training Cost.}

\begin{figure}[t]
  \centering
  \includegraphics[width=0.95\linewidth]{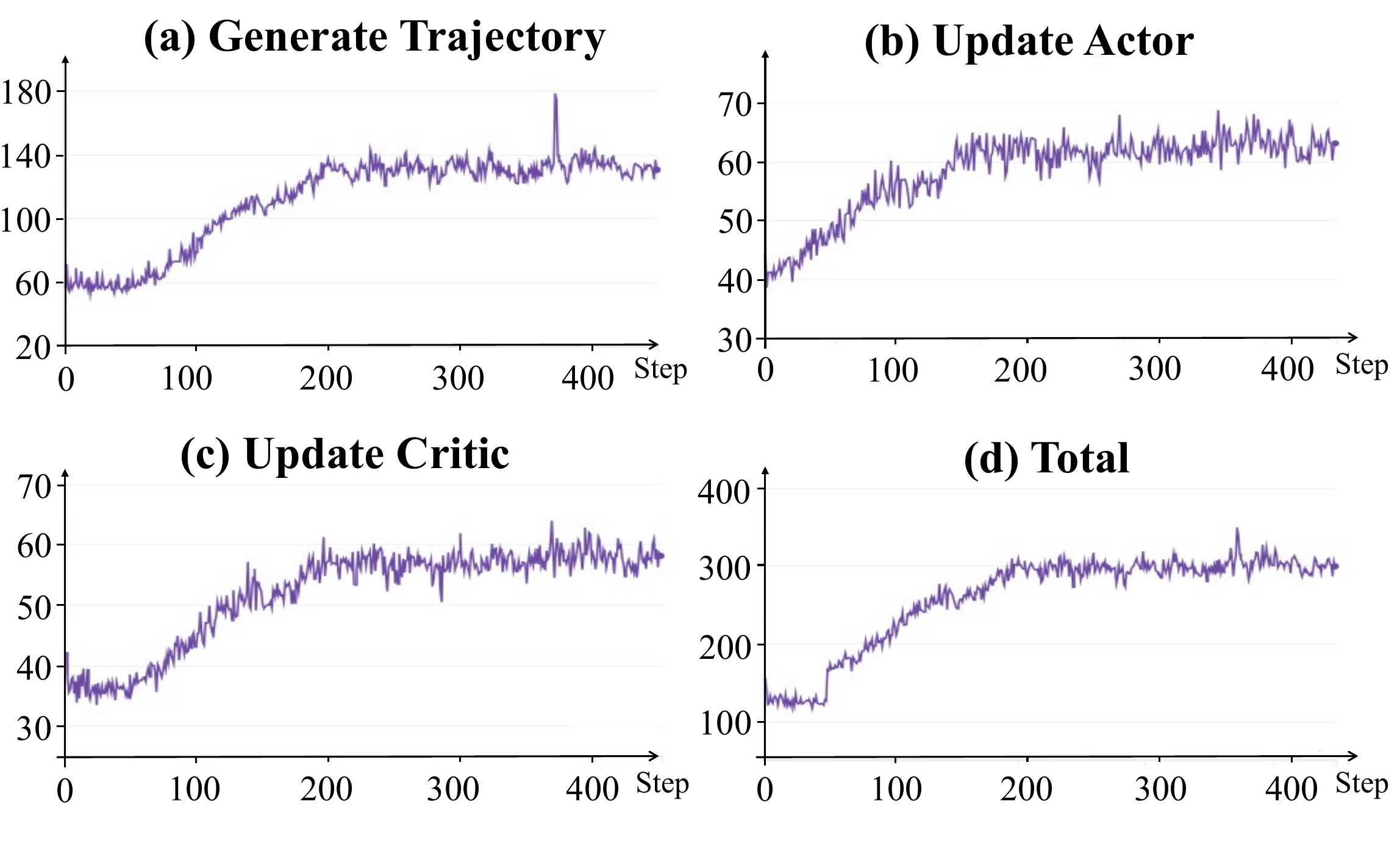}
  \caption{Per-step training time consumption (in seconds) of PaperScout. (a) Rollout phase; (b) Actor update; (c) Critic  update; (d) Total time consumption.
}
\label{fig:train_time}
\end{figure}

PSPO training is conducted on four NVIDIA H800 (80GB) GPUs, consisting of 100 critic pretraining steps followed by 200 joint actor–critic optimization steps, with a total training time of approximately 12.6 hours (≈4,000s for critic pretraining and ≈39,750s for PSPO training). As shown in Figure \ref{fig:train_time}, after the warm-up phase the per-step training time stabilizes at about 290–310 seconds. Among the components, rollout dominates the cost due to multi-turn interaction with the retrieval environment, requiring approximately 135–145 seconds per step, while actor and critic updates take about 60–65 seconds and 55–60 seconds, respectively. Overall, policy optimization accounts for less than half of the step time, indicating that most computational overhead arises from environment interaction rather than policy updates.

\paragraph{Inference Efficiency.}

% Beyond training cost, we further evaluate the interaction efficiency of the agent by analyzing the recall–time trade-off. Figure 2 shows the recall growth curves of different systems on the RealScholarQuery dataset. Qwen3-4B-PSPO demonstrates consistently faster recall improvement than other agent-based methods, including Qwen3-Max, PaSa, and SPAR. The model quickly reaches competitive recall within the first few seconds and continues to improve steadily during interaction, ultimately surpassing these baselines in final performance.

% Moreover, the recall–time curve of Qwen3-4B-PSPO remains competitive with traditional retrieval systems such as Google Scholar and Google Search. Under comparable time budgets, the proposed system matches or even exceeds their recall performance. These results indicate that PSPO improves the efficiency of multi-turn retrieval interactions while maintaining strong retrieval effectiveness relative to conventional search pipelines.

\begin{figure}[t]
  \centering
  \includegraphics[width=0.95\linewidth]{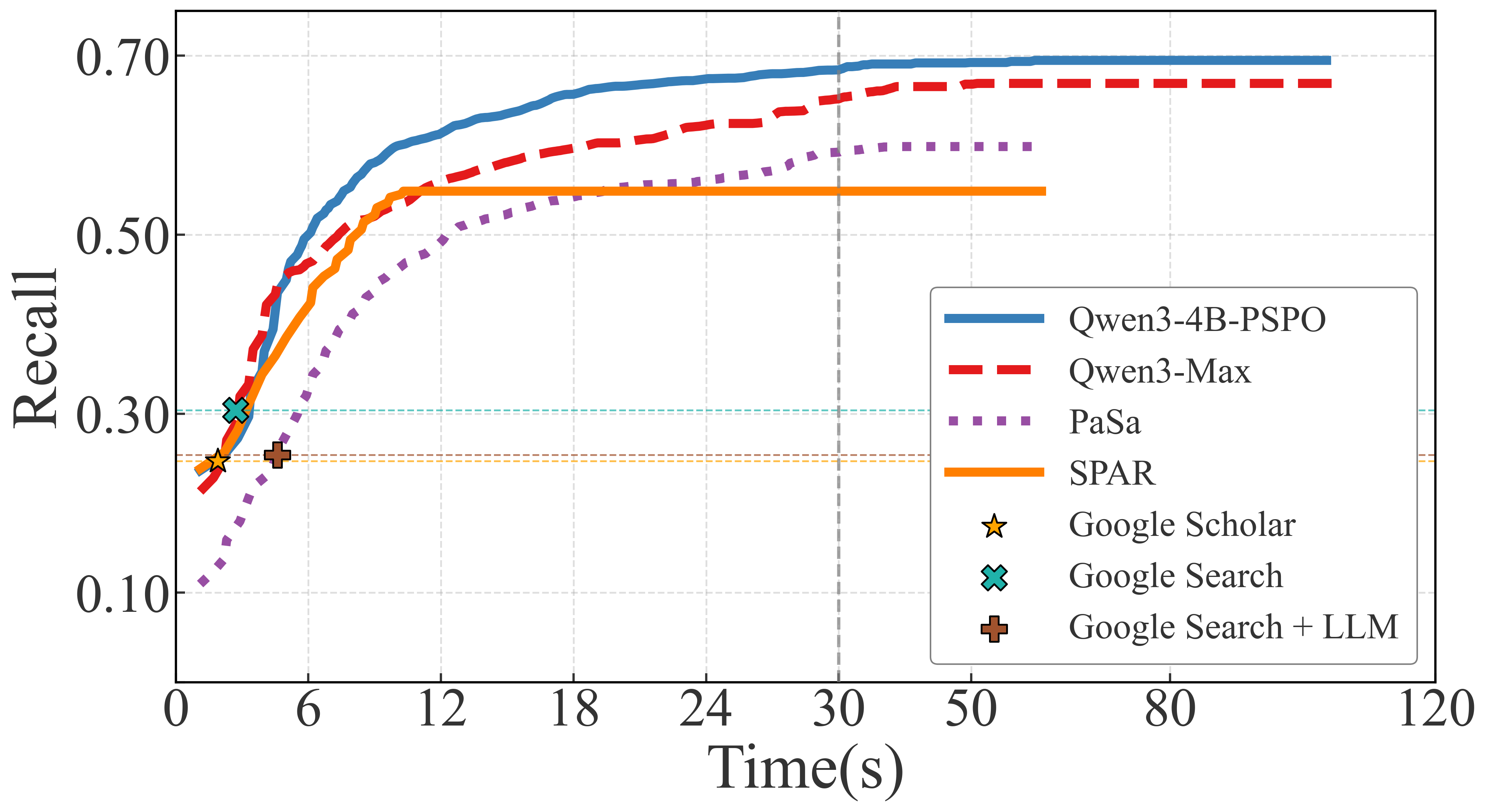}
  \caption{Recall–time curves on RealScholarQuery. PaperScout reaches near-optimal recall within 30 seconds, outperforming both agent-based and traditional retrieval baselines over runtime.
}
\label{fig:recall_time}
\end{figure}

Beyond training cost, we also examine whether the learned policy leads to efficient inference behavior. Figure \ref{fig:recall_time} presents recall as a function of runtime on RealScholarQuery, reflecting how much retrieval gain each system obtains per unit time. Compared with other agent-based baselines, including Qwen3-Max, PaSa, and SPAR, Qwen3-4B-PSPO achieves faster recall improvement and higher final recall under similar or shorter time budgets, showing that PSPO improves not only final effectiveness but also the efficiency of the retrieval process.

Moreover, despite requiring multi-turn reasoning and repeated tool interaction, Qwen3-4B-PSPO remains competitive with conventional retrieval systems such as Google Scholar and Google Search in terms of recall under comparable runtime budgets. This indicates that the additional interaction cost introduced by the agent framework is effectively offset by more informed and productive search decisions. In other words, while PSPO incurs a moderate one-time training cost, it yields a policy with substantially improved inference efficiency, leading to a better effectiveness–efficiency trade-off overall.

% \newpage

\section{Discussion}

\paragraph{Distinction from General Search Agents.} PaperScout differs from general search or deep research agents such as Search-R1 \cite{jin2025searchr} and WebThinker \cite{li2026webthinker} in task structure and optimization objective. While these agents usually retrieve information to support a final textual answer or report, academic paper search takes the retrieved paper set itself as the target output. The task therefore directly emphasizes relevance, coverage, and ranking quality, and naturally provides verifiable process feedback: each Search or Expand action updates the paper pool and yields measurable marginal retrieval gains. This structure makes paper search particularly suitable for multi-turn decision-making, process-level rewards, and interaction-level optimization.

\paragraph{Scope of Sequence-Level Optimization.}
PSPO assumes that each complete model response forms a meaningful atomic action,
which holds in PaperScout because a response directly triggers tool execution
and receives a step reward. For agents whose responses mix multiple decisions
or long-form generation, sequence-level credit may again become coarse,
motivating hierarchical or segment-level extensions.

\section{Conclusion}
In this paper, we present PaperScout, an autonomous agent that formulates academic paper search as a sequential decision-making process. By explicitly deciding when and how to invoke \texttt{search} and \texttt{expand} tools based on accumulated retrieval context, PaperScout moves beyond static workflows and enables flexible multi-turn paper exploration. We further introduce Proximal Sequence Policy Optimization (PSPO), a process-aware, sequence-level policy optimization method that aligns optimization with agent--environment interaction and enables stable training of multi-turn retrieval agents. Experiments on both synthetic
and real-world benchmarks demonstrate that PaperScout trained by PSPO achieves superior retrieval performance compared to existing methods.

% \begin{landscape}

\clearpage

\section*{Limitations}
Despite the effectiveness of our approach in multi-turn paper search, several limitations remain. 
First, our current evaluation primarily focuses on the computer science domain, and extending the study to broader research areas would further validate the generality of the proposed framework. 
Second, PaperScout relies on papers that are publicly accessible through online search engines and open repositories; retrieving papers from restricted or paywalled sources remains challenging and may limit coverage in certain domains. 
Third, our current implementation adopts a relatively limited search backend, and incorporating multi-source retrieval from heterogeneous scholarly databases could further improve robustness and recall. 
Finally, citation expansion currently considers only outgoing references, while leveraging incoming citations and richer citation graph signals may provide additional retrieval cues. 
We leave these directions for future work.

\section*{Acknowledgments}

This work was supported by grants from the National Natural Science Foundation of China (62502486), the Fundamental Research Funds for the Central Universities of China, USTC Kunpeng-Ascend Scientific and Education Innovation Excellence Center.

\bibliography{PaperScout}

% \newpage

\appendix

\section{Experimental Details}

\paragraph{Implementation of PaperScout.}
We conduct reinforcement learning for PaperScout on four NVIDIA H800 GPUs, using Qwen3-4B-Instruct-2507\footnote{https://www.modelscope.cn/models/Qwen/Qwen3-4B-Instruct-2507} as the backbone model. We set the actor and critic learning rates to $1\times10^{-6}$ and $1\times10^{-5}$, respectively, with a per-gpu batch size of 4. For reward computation, we set the top-$k$ to 3 to reduce reward variance across queries with varying difficulty. We further set the score threshold $\tau=0.01$ and repeated penalty $\eta=0.5$. To stabilize training, we freeze the actor and pre-train the critic for 100 steps before joint optimization. At each turn, the agent receives up to 10 unexpanded and 10 expanded papers. If the paper pool remains unchanged for three turns, the retrieval terminates. More training details are shown in Table \ref{tab:train_details}.

% \subsection{Dataset Description}
% Our experiments are conducted on two benchmarks, AutoScholarQuery and RealScholarQuery \cite{he2025pasa}, which represent synthetic and expert-curated scholarly search scenarios. AutoScholarQuery is a large-scale synthetic benchmark constructed from recent top-tier conference papers, including ICLR, ICML, NeurIPS, ACL, and CVPR, while RealScholarQuery is a human-annotated benchmark designed to evaluate realistic and challenging scholarly search settings. Detailed statistics of the datasets are summarized in Table~\ref{tab:datasets}.

% Our experiments are conducted on two benchmarks, AutoScholarQuery and RealScholarQuery \cite{he2025pasa}, which represent synthetic and expert-curated scholarly search scenarios. Detailed statistics of the two datasets are summarized in Table~\ref{tab:datasets}.

% \begin{itemize}
%     \item \textbf{AutoScholarQuery}: A large-scale synthetic benchmark constructed from recent top-tier conference papers, including ICLR, ICML, NeurIPS, and ACL.
%     \item \textbf{RealScholarQuery}: A human-annotated benchmark designed to evaluate realistic and challenging scholarly search scenarios.
% \end{itemize}

% \begin{table}[ht]
%     \centering
%     \begin{tabular}{p{0.45\linewidth}ccc}
%         \toprule
%         \textbf{Dataset} & \textbf{\#Train} & \textbf{\#Dev} & \textbf{\#Test} \\
%         \midrule
%         AutoScholarQuery & 33551 & 1000 & 1000 \\
%         RealScholarQuery & -- & -- & 50 \\
%         \bottomrule
%     \end{tabular}
%     \caption{Details of the datasets}
%     \label{tab:datasets}
% \end{table}

\label{sec:imple_detials}

% \begin{table}[ht]
% \centering
% \small
% \begin{tabular}{lc}
% \toprule
% Name & Value \\
% \midrule
% Backbone & Qwen3-4B-Instruct-2507 \\
% Advantage estimator & GAE \\
% Actor learning rate & $1\times10^{-6}$ \\
% Critic learning rate & $1\times10^{-5}$ \\
% Train batch size & 128 \\
% Actor mini-batch size & 320 \\
% Actor micro-batch size & 4 \\
% Critic mini-batch size & 128 \\
% Critic micro-batch size & 8 \\
% Max prompt length & 10,240 \\
% Max response length & 4,096 \\
% KL coefficient & 0.001 \\
% Clip ratio (low) & $3\times10^{-4}$ \\
% Clip ratio (high) & $4\times10^{-4}$ \\
% Training steps & 300 \\
% \bottomrule
% \end{tabular}
% \caption{Hyperparameters used in PSPO training.}
% \end{table}

\begin{table}[ht]
\centering
\small
\setlength{\tabcolsep}{5pt}

\resizebox{1.0\linewidth}{!}{
\begin{tabular}{lll}

\toprule
Category & Hyperparameter & Value \\
\midrule
\multirow{4}{*}{Optimization}
 & Actor learning rate & $1\times10^{-6}$ \\
 & Critic learning rate & $1\times10^{-5}$ \\
 & Policy freezing steps & 100 \\
 & Total training steps & 300 \\

\midrule
\multirow{5}{*}{RL Hyperparameters}
 & Advantage estimator & GAE \\
 & Discount factor $\gamma$ & 0.99 \\
 & GAE parameter $\lambda$ & 0.95 \\
 & Clip ratio $(\varepsilon_{low})$ & $3\times10^{-4}$ \\
 & Clip ratio $(\varepsilon_{high})$ & $4\times10^{-4}$ \\
 & KL coefficient & 0.001 \\

\midrule
\multirow{4}{*}{Batching}
 & Train batch size & 128 \\
 & Actor mini-batch size & 320 \\
 & Critic mini-batch size & 128 \\
 & Actor micro-batch size & 4 \\
 & Critic micro-batch size & 8 \\

\midrule
\multirow{2}{*}{Sequence Length}
 & Max prompt length & 10,240 \\
 & Max response length & 4,096 \\

\bottomrule
\end{tabular}
}
\caption{Hyperparameters used for PSPO training.}
\label{tab:train_details}
\end{table}

\paragraph{Scorer.} Given a user query together with the title and abstract of a paper, the scorer evaluates their semantic relevance and produces a continuous score that guides the retrieval decision of PaperScout. Specifically, the scorer prompts a language model to perform a relevance judgment, and defines the relevance score as the probability of generating the ``True" token in the model output, yielding a value in $[0,1]$. For this component, we utilize the off-the-shelf pasa-7b-selector\footnote{https://www.modelscope.cn/models/bytedance-research/pasa-7b-selector}, a Qwen2.5-7B based model optimized for relevance assessment. The detailed prompt is shown in Appendix \ref{prompt:selector}.

\section{Details about LLM-score}
\label{sec:llm_score}

\paragraph{Evaluation Protocol.}
Recently, numerous studies have explored the evaluation of large language model capabilities \cite{liu2025elo,liu2026fewer}. In this work, we employ an LLM-as-a-Judge to assess the relevance between a user query and a candidate paper based on its title and abstract, using three independent evaluators—DeepSeek-V3.2, Qwen3-Max, and GPT-5.1. To ensure reliable and unbiased judgments, the evaluators are explicitly instructed to focus solely on semantic relevance, while ignoring factors such as writing quality, length, style, or popularity. Relevance is rated on a discrete four-level scale from 0 to 3, with clearly defined semantic criteria for each level, which helps constrain the models’ decision boundaries and reduce ambiguity in scoring. Evaluators are instructed to return relevance scores in a strict JSON format to ensure consistent and structured outputs for downstream analysis. The detailed prompt is provided in Appendix \ref{prompt:llm_score}.

\paragraph{Inter-model Consistency.} We randomly sample 500 query–paper pairs from each method’s test results, yielding a total of 3,000 samples, and analyze the consistency of the three LLM-based relevance scores for each sample. For each query–paper pair, we compute the standard deviation (STD) of the three scores as well as their maximum score gap (Max-Gap) to quantify inter-model agreement. As shown in Table \ref{tab:llm-score-consistency}, most samples exhibit strong consistency: 2,629 cases receive identical scores from all three models. The remaining discrepancies are predominantly minor, with only small score differences, while larger disagreements are rare and no samples exhibit extreme conflicts. These results indicate that the LLM-based scoring provides a stable and reliable signal for relevance evaluation.

\begin{table}[ht]
    \centering
    \resizebox{0.9\linewidth}{!}{
        \begin{tabular}{l c c c c c}
            \toprule
            STD &
            0.00 & 0.22 & 0.66 & 0.88 & $>0.88$ \\
            Max-Gap & 0    & 1   & 2  & 2  & 3 \\
            \midrule
            Count   & 2,629 & 336 & 25 & 10 & 0 \\
            \bottomrule
        \end{tabular}
    }
    \caption{Consistency analysis of LLM-based relevance scores across three evaluators.}
    \label{tab:llm-score-consistency}
\end{table}

\vspace{-0.1in}

\begin{figure*}[t]
  \centering
  \includegraphics[width=\linewidth]{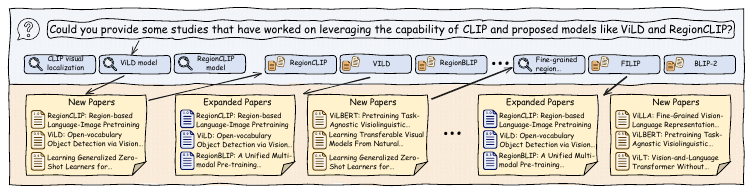}
  \caption{Case study of PaperScout. The agent alternates between \texttt{Search} and \texttt{Expand} based on accumulated retrieval context, re-issuing \texttt{Search} when \texttt{Expand} yields diminishing returns to introduce new directions.}

  \label{fig:case}
\end{figure*}

\paragraph{Agreement with Human Judgment.}

To assess whether LLM-based relevance scoring is aligned with human judgment, we further compare LLM-assigned scores with human annotations. Specifically, we randomly sample QA-pairs  and compare their relevance scores on the same four-level scale. To ensure comparability, human annotators are instructed to follow the same scoring criteria as the LLM evaluators, as specified in Scoring Rubric \ref{prompt:llm_score}.

% \begin{table}[ht]
%     \centering
%     \begin{tabular}{c|cccc|c}
%         \toprule
%         \multirow{2}{*}{\textbf{LLM Score}} 
%         & \multicolumn{4}{c|}{\textbf{Human Score}} 
%         & \multirow{2}{*}{\textbf{Row Total}} \\
%         \cmidrule(lr){2-5}
%         & \textbf{0} & \textbf{1} & \textbf{2} & \textbf{3} & \\
%         \midrule
%         0 & \textbf{88} & 11 & 1  & 0  & 100 \\
%         1 & 0  & \textbf{82} & 15 & 3  & 100 \\
%         2 & 0  & 9  & \textbf{87} & 5  & 100 \\
%         3 & 0  & 1  & 5  & \textbf{94} & 100 \\
%         \bottomrule
%     \end{tabular}
%     \caption{Distribution of LLM-based versus human relevance scores. Bold numbers indicate exact agreement.}
%     \label{tab:gpt_human_score_consistency}
% \end{table}

\begin{table}[t]
    \centering
    \resizebox{0.95\linewidth}{!}{
    \begin{tabular}{c|cccc|c}
        \toprule
        \multirow{2}{*}{\textbf{LLM Score}} 
        & \multicolumn{4}{c|}{\textbf{Human Score}} 
        & \multirow{2}{*}{\textbf{Row Total}} \\
        \cmidrule(lr){2-5}
        & \textbf{0} & \textbf{1} & \textbf{2} & \textbf{3} & \\
        \midrule
        0 & \textbf{88} & 11 & 1  & 0  & 100 \\
        1 & 0  & \textbf{82} & 15 & 3  & 100 \\
        2 & 0  & 8  & \textbf{87} & 5  & 100 \\
        3 & 0  & 1  & 5  & \textbf{94} & 100 \\
        \midrule
        Col Total & 88 & 102 & 108 & 102 & 400 \\
        \bottomrule
    \end{tabular}
    }
    \caption{Distribution of LLM-based versus human relevance scores. Bold numbers indicate exact agreement.}
    \label{tab:gpt_human_score_consistency}
\end{table}

As shown in Table~\ref{tab:gpt_human_score_consistency}, LLM-based scores are highly consistent with human annotations, with most samples falling on the diagonal. The agreement is especially strong for clear cases: 88\% of samples assigned a score of `0' by LLMs are also labeled as `0' by humans, and 94\% of those assigned `3' are also labeled as `3'. Most remaining discrepancies occur between adjacent scores, mainly around intermediate levels such as `1' and `2', suggesting that disagreements largely come from borderline cases rather than severe misjudgments. Overall, LLM-based scoring is well aligned with human judgment and provides a reliable relevance signal for our evaluation.

\section{Case Study}

\subsection{Case: Adaptive Tool-call Behavior}

\label{sec:case_d1}
Figure \ref{fig:case} presents a representative case illustrating PaperScout’s multi-turn retrieval behavior on a complex query involving CLIP, ViLD, and RegionCLIP. At the initial stage, the agent decomposes query into multiple semantic facets and issues parallel \texttt{Search} tool calls targeting different aspects of vision–language models. Based on the retrieved seed papers, the agent performs several rounds of \texttt{Expand} tool calls, progressively exploring region-level and fine-grained vision–language pretraining methods with strong semantic coherence.

Notably, after expansion along existing directions becomes saturated, the agent initiates a new \texttt{Search} call in a later turn, opening a previously unexplored research direction, fine-grained region understanding. This re-search step moves the agent beyond local expansion by introducing fresh candidate papers, which are further refined through subsequent \texttt{Expand} calls. Such behavior demonstrates that PaperScout does not follow a fixed search–then–expand pipeline, but instead automatically alternates between \texttt{Search} and \texttt{Expand} based on accumulated context, constructing a structured and evolving paper exploration trajectory.

\subsection{Case: Comparison with Other Agents}

% To further understand the gains of PaperScout over structured multi-turn baselines such as PaSa and SPAR, we report its tool-use analysis and retrieval performance on representative RealScholarQuery cases in Table~\ref{tab:paperscout_tool_traces}. Unlike the specific case discussed in above, this analysis highlights common patterns across query types. PaperScout shows larger advantages on enumerative and multi-constraint queries, where relevant papers are often scattered across different terminologies, subfields and citation neighborhoods.

To further understand the gains of PaperScout over structured multi-turn baselines such as PaSa and SPAR, we summarize its tool-use patterns and retrieval performance on representative RealScholarQuery cases in Table~\ref{tab:paperscout_tool_traces}. Beyond the single case discussed above, this analysis reveals common patterns across query types. PaperScout shows larger advantages on enumerative and multi-constraint queries, where relevant papers are often scattered across different terminologies, subfields, and citation neighborhoods.

\begin{table*}[t]
\centering
\scriptsize
\setlength{\tabcolsep}{4pt}
\renewcommand{\arraystretch}{1.18}
\begin{tabular}{@{}
>{\RaggedRight\arraybackslash}p{0.33\textwidth}
>{\RaggedRight\arraybackslash}p{0.30\textwidth}
>{\RaggedRight\arraybackslash}p{0.33\textwidth}
@{}
}
\toprule
\textbf{Query}
& \textbf{Details}
& \textbf{Tool-use Strategy of PaperScout} \\
\midrule

Help me search for the work related to the synthetic data of large language models. I want to know how to automatically generate large-scale, high-quality, diverse, difficult, and valuable long thought data for learning.
&
\textbf{Action/ Search/ Expand cnt. (@PaperScout):}\par
15 / 60 / 10
\par\vspace{2pt}
\textbf{Recall@all (@PaperScout/ PaSa/ SPAR):}\par
1.0000 / 0.0000 / 0.0000
&
First searches from multiple angles, including synthetic data, long-thought data, and high-quality, diverse, and difficult data. After finding \texttt{DeepSeek-Prover}, it expands related seed papers and recovers \texttt{MUSTARD}. \\

\midrule

I am looking for research papers on the construction of multimodal foundation models that support both visual and audio inputs. These models should be pre-trained on large-scale datasets, including visual, audio, and audio-visual data. Please exclude survey papers.
&
\textbf{Action/ Search/ Expand cnt. (@PaperScout):}\par
9 / 19 / 20
\par\vspace{2pt}
\textbf{Recall@all (@PaperScout/ PaSa/ SPAR):}\par
0.7381 / 0.6667 / 0.4524
&
Starts with broad multimodal and audio-visual foundation-model searches, then heavily expands representative papers, which fits high-recall, multi-constraint queries. \\

\midrule

Give me papers about how to rank search results by the use of LLM.
&
\textbf{Action/ Search/ Expand cnt. (@PaperScout):}\par
14 / 46 / 12
\par\vspace{2pt}
\textbf{Recall@all (@PaperScout/ PaSa/ SPAR):}\par
0.6410 / 0.4359 / 0.3846
&
Searches LLM ranking and reranking first, then expands core papers on listwise ranking and zero-shot rankers. Later searches are refined toward fine-tuning, self-consistency, setwise prompting, and ranking losses. \\

\midrule

Show me research on the long video description. Here, long videos are defined as those with a duration of at least several minutes.
&
\textbf{Action/ Search/ Expand cnt. (@PaperScout):}\par
15 / 52 / 10
\par\vspace{2pt}
\textbf{Recall@all (@PaperScout/ PaSa/ SPAR):}\par
0.6286 / 0.5143 / 0.3429
&
Begins with long video description and multi-minute video queries, then refines toward temporal understanding, hierarchical event modeling, episodic memory, scene graphs, and long-form video summarization. \\

\midrule

Provide papers on methods that protect the generation quality of LLMs under vocabulary watermarking settings.
&
\textbf{Action/ Search/ Expand cnt. (@PaperScout):}\par
11 / 28 / 21
\par\vspace{2pt}
\textbf{Recall@all (@PaperScout/ PaSa/ SPAR):}\par
0.7297 / 0.5676 / 0.5676
&
Uses many Expand calls around core watermarking papers, then supplements them with searches for semantic-aware, adaptive, undetectable, and robust watermarking under quality-preservation constraints. \\

\midrule

Give me all visual-LLM models that are MoE architecture.
&
\textbf{Action/ Search/ Expand cnt. (@PaperScout):}\par
14 / 40 / 20
\par\vspace{2pt}
\textbf{Recall@all (@PaperScout/ PaSa/ SPAR):}\par
0.8500 / 0.8000 / 0.5500
&
Starts with visual-LLM and MoE architecture searches, expands VLM/MoE seed papers, then explores application subdomains such as medical imaging, robotics perception, autonomous driving, and real-world scene understanding. \\

\midrule

Papers that propose methods based on large language models and evaluate their performance through experiments on the HotPotQA dataset.
&
\textbf{Action/ Search/ Expand cnt. (@PaperScout):}\par
18 / 63 / 12
\par\vspace{2pt}
\textbf{Recall@all (@PaperScout/ PaSa/ SPAR):}\par
0.5417 / 0.4167 / 0.4583
&
Repeatedly rewrites the query around HotPotQA, multi-hop reasoning, retrieval, chain-of-thought, external knowledge, performance metrics, and error analysis, with occasional Expand calls on relevant seed papers. \\

\bottomrule
\end{tabular}
\caption{
Tool-use analysis and retrieval performance on representative RealScholarQuery cases, where PaperScout shows clear advantages over PaSa and SPAR.
}
\label{tab:paperscout_tool_traces}
\end{table*}

This pattern stems from the agents' different retrieval strategies. PaSa learns search, expansion, and stop actions, but still follows a paper-queue traversal protocol: it first collects candidates and then expands their citation neighborhoods. SPAR improves retrieval through a structured modular pipeline, mainly strengthening query understanding and retrieval enhancement. By contrast, PaperScout dynamically controls retrieval based on the evolving candidate pool. Instead of following a prescribed traversal pattern, it can continue searching, expand promising seeds, or reformulate queries when current results are insufficient. This flexibility helps explain its higher Recall@all: PaperScout decomposes constrained queries into searchable facets and progressively explores sub-domains, making it well suited to comprehensive academic paper discovery.

\section{Full Prompt Set}

\begin{tcolorbox}[breakable, title=Paper Search Agent Prompt,colback=gray!5,colframe=nmgray!75!black,before skip=1pt, after skip=1pt,fontupper=\linespread{0.8}\selectfont]
{\footnotesize
\textbf{System}

You are a research agent. Your goal is to find papers relevant to the User Query.

\textbf{User Query} 

(...Detailed User Query...)

\textbf{History Actions}

(...Detailed History Actions...)

\textbf{Paper List}

(...Detailed Paper List...)

\textbf{Instructions}
\begin{itemize}
    \item Analyze the Paper List and History Actions to determine the next set of actions. Enclose your analysis of the state and decision logic within \texttt{<analysis>...</analysis>} tags.
    \item You support parallel tool calling. You should output multiple tool calls in a single step if several independent actions are valuable at the current state.
    \item Attend to the history actions and avoid expanding the same papers.
\end{itemize}

\textbf{Output Format}
\begin{PromptCode}
<analysis>
[Your analysis of the current state and decision logic...]
</analysis>
<tool_call>
[Tool call 1]
</tool_call>
<tool_call>
[Tool call 2]
</tool_call>
...
\end{PromptCode}

\textbf{Tool Schema}
\begin{PromptCode}
SEARCH_TOOL_SCHEMA = {
    "type": "function",
    "function": {
        "name": "search",
        "description": (
            "Search for relevant papers in the arXiv repository."
        ),
        "parameters": {
            "type": "object",
            "properties": {
                "query": {
                    "type": "string",
                    "description": (
                        "A single search query (natural language or keywords). "
                        "No field scopes (ti:/abs:) or boolean ops. Must differ from all 
                        history queries."
                    ),
                }
            },
            "required": ["query"],
        },
    },
}

EXPAND_TOOL_SCHEMA = {
    "type": "function",
    "function": {
        "name": "expand",
        "description": (
            "Expand from an existing paper by following its references to surface additional 
            relevant works. "
            "Use this when search is saturated and you want to broaden coverage around a known 
            paper."
        ),
        "parameters": {
            "type": "object",
            "properties": {
                "arxiv_id": {
                    "type": "string",
                    "description": (
                        "The arXiv identifier (e.g., '1706.03762') of a paper already in the 
                        current paper list."
                    ),
                }
            },
            "required": ["arxiv_id"],
        },
    },
}

PAPERSEARCH_TOOL_SCHEMAS = [SEARCH_TOOL_SCHEMA, EXPAND_TOOL_SCHEMA]
\end{PromptCode}

}
\end{tcolorbox}

~\\

\label{prompt:selector}
\begin{tcolorbox}[breakable, title=Scorer Prompt,colback=gray!5,colframe=nmgray!75!black,before skip=1pt, after skip=1pt,fontupper=\linespread{0.8}\selectfont]
{\footnotesize
\textbf{System}

You are an elite researcher in the field of AI, conducting research on 

(...Detailed User Query...)

Evaluate whether the following paper fully satisfies the detailed requirements of the user query and provide your reasoning. Ensure that your decision and reasoning are consistent.

\textbf{Searched Paper}
\begin{itemize}
    \item Title: (...Detailed Title...)
    \item Abstract: (...Detailed Abstract...)
    \item User Query: (...Detailed User Query...)
\end{itemize}

\textbf{Output format}
\begin{itemize}
    \item Decision: True~/~False
    \item Reason: ... 
\end{itemize}
}
\end{tcolorbox}

~\\

\label{prompt:llm_score}
\begin{tcolorbox}[breakable, title=LLM Judge Prompt,colback=gray!5,colframe=nmgray!75!black,before skip=1pt, after skip=1pt,fontupper=\linespread{0.8}\selectfont]
{\footnotesize
\textbf{System}

You are an expert LLM-as-a-Judge for academic paper search.
Your task is to evaluate how relevant a candidate paper is to a given research query.
You must strictly follow the scoring rubric and output only the required JSON.
You should ignore writing quality, length, style, or popularity.
Only evaluate semantic relevance between the query and the paper content.

\textbf{Query} 

(...Detailed User Query...)

\textbf{Paper Candidate}

\begin{itemize}
    \item Title: 
    \item Abstract: 
\end{itemize}

\textbf{Scoring Rubric (relevance\_level)}

\begin{itemize}
    \item - 3 = Highly relevant — Directly addresses the core research question or problem expressed in the query.
    \item - 2 = Moderately relevant — About a closely related topic, but only partially matches the intent or focuses on a sub-aspect.
    \item - 1 = Weakly relevant — Belongs to the broader area but only tangentially connected to the query.
    \item - 0 = Irrelevant — Not relevant to the query in any meaningful way.
\end{itemize}

\textbf{Output Format (must be valid JSON)}

\begin{PromptCode}
{
  "relevance_level": 0/1/2/3,
  "confidence": "low/medium/high",
  "reason_short": "A brief one-sentence justification"
}
\end{PromptCode}

You MUST output valid JSON and nothing else.

}
\end{tcolorbox}

\section{An Example for PaperScout}

\begin{tcolorbox}[breakable, title=Query ,colback=gray!5,colframe=nmgray!75!black,before skip=1pt, after skip=1pt,fontupper=\linespread{0.8}\selectfont]
% {\footnotesize
Could you provide some studies that have worked on leveraging the capability of CLIP and proposed models like ViLD and RegionCLIP?
% }
\end{tcolorbox}

\begin{tcolorbox}[breakable, title=Step 1 ,colback=gray!5,colframe=nmgray!75!black,before skip=1pt, after skip=1pt,fontupper=\linespread{0.8}\selectfont]
{\footnotesize
\textbf{Paper List:}

Empty list

\textbf{Analysis:}

The initial paper list is empty, and there are no history actions to avoid. The user is interested in studies that leverage CLIP's capabilities, specifically referencing models like ViLD and RegionCLIP. Since there are no existing papers in the pool, we must first search for relevant papers that discuss CLIP-based models or specifically mention ViLD and RegionCLIP. To ensure comprehensive coverage, we should perform a broad search query with keywords such as ``CLIP", ``ViLD", and ``RegionCLIP" to identify foundational and related works.
Given that the user is interested in the application and development of CLIP-based models like ViLD and RegionCLIP, initiating multiple search queries with slightly different but complementary keywords (e.g., ``CLIP-based visual localization", ``ViLD model", ``RegionCLIP model") will help capture a diverse range of relevant papers. This ensures that we cover both general CLIP applications and the specific models mentioned in the query.
We will perform three initial search queries:
``CLIP visual localization"
``ViLD model"
``RegionCLIP model"
These queries will yield a set of foundational and relevant papers directly matching the user's interests, enabling us to form a robust initial pool to proceed with further expansion or refinement. 

\textbf{Tool Call:}
\begin{PromptCode}
{"name": "search", "arguments": {"query": "CLIP visual localization"}}
{"name": "search", "arguments": {"query": "ViLD model"}}
{"name": "search", "arguments": {"query": "RegionCLIP model"}}
\end{PromptCode}

}
\end{tcolorbox}

\begin{tcolorbox}[breakable, title=Step 2 ,colback=gray!5,colframe=nmgray!75!black,before skip=1pt, after skip=1pt,fontupper=\linespread{0.8}\selectfont]
{\footnotesize
\textbf{Paper List:}

Paper Pool Status:

[\textcolor{green}{EXP}]: Paper has been expanded (already used as a seed for more papers).

[\textcolor{red}{NEW}]: New paper found via search or expansion, candidate for further exploration.

Format: [arxiv\_id] (score) [STATUS] Title

[\textcolor{orange}{2112.09106}] (\textcolor{blue}{1.00}) [\textcolor{red}{NEW}] RegionCLIP: Region-based Language-Image Pretraining 

Abstract: Contrastive language-image pretraining (CLIP) performs poorly on region-level recognition due to a domain shift between whole-image and region-text alignment.
We propose RegionCLIP to learn region-level visual representations aligned with textual concepts, significantly improving open-vocabulary object detection performance.
...

[\textcolor{orange}{2104.13921}] (\textcolor{blue}{0.99}) [\textcolor{red}{NEW}] Open-vocabulary Object Detection via Vision and Language Knowledge Distillation 

Abstract: We propose ViLD, a vision-language knowledge distillation framework that transfers open-vocabulary knowledge from a pretrained image-text model to a two-stage detector.
ViLD achieves state-of-the-art performance on LVIS and generalizes well to COCO, PASCAL VOC, and Objects365 without finetuning.
...

[\textcolor{orange}{2302.00275}] (\textcolor{blue}{0.40}) [\textcolor{red}{NEW}] Learning Generalized Zero-Shot Learners for Open-Domain Image Geolocalization 

Abstract: We introduce StreetCLIP, a foundation model that achieves state-of-the-art zero-shot performance for open-domain image geolocalization.
By meta-learning with synthetic captions, our method effectively transfers CLIP’s generalized zero-shot capability to the geolocalization domain.
...

(...More Papers...)

[\textcolor{orange}{2305.05095}] (\textcolor{blue}{0.12}) [\textcolor{red}{NEW}] Less is More: Removing Text-regions Improves CLIP Training Efficiency and Robustness 

Abstract: We show that removing images with text regions improves both the efficiency and robustness of CLIP training.
The proposed filtering strategy boosts classification and retrieval accuracy and defends against typographic adversarial attacks.
...

\textbf{Analysis:}

The user query includes the topic about `RegionCLIP model'' and already performed a search for related topics, including `CLIP visual localization'', `ViLD model'', and `RegionCLIP model''. The current paper list includes several relevant works, with the most directly relevant being [2112.09106] (RegionCLIP: Region-based Language-Image Pretraining), which is the primary paper matching the exact query.

From the list:
[2112.09106] is the core paper on RegionCLIP and directly addresses the user's initial interest.
[2104.13921] (ViLD) is closely related, as it deals with open-vocabulary object detection via vision and language distillation—complementary to RegionCLIP's goal of fine-grained region understanding.
[2308.02299] (RegionBLIP) is also relevant as it extends multimodal models to regional comprehension, implying a similar focus on region-level understanding.
[2310.18340] (UrbanCLIP) is less directly relevant, as it applies CLIP to urban region profiling rather than object detection or region alignment.
[2302.00275] and [2309.16020] focus on geolocation, which is tangentially related via vision-language alignment but not directly on region-level features.

We should expand beyond the initial RegionCLIP paper to explore related works that enhance region-level understanding, especially those that build upon or extend CLIP for regional visual representations. Given that [2112.09106] is the primary paper, we should expand it to surface more foundational or downstream works that might be relevant.

Moreover, [2104.13921] and [2308.02299] are strong secondary candidates because they also involve region-image-text alignment in a multimodal setting. However, they were not previously expanded, so expanding these papers would provide a more robust understanding of the ecosystem.

Therefore, the next actions should be:
Expand the primary paper [2112.09106] to uncover related foundational or downstream works.
Expand [2104.13921] (ViLD) to explore its references, as it shares a similar vision (open-vocabulary detection via language vision alignment).
Expand [2308.02299] (RegionBLIP) to discover related works on regional comprehension in multimodal models.

These expansions will enrich the paper pool without repeating prior searches or expanding redundant papers (e.g., earlier ones not in the current list).

\textbf{Tool Call:}
\begin{PromptCode}
{"name": "expand", "arguments": {"arxiv_id": "2112.09106"}}
{"name": "expand", "arguments": {"arxiv_id": "2104.13921"}}
{"name": "expand", "arguments": {"arxiv_id": "2308.02299"}}
\end{PromptCode}
}
\end{tcolorbox}

\begin{tcolorbox}[breakable, title=Step 3 ,colback=gray!5,colframe=nmgray!75!black,before skip=1pt, after skip=1pt,fontupper=\linespread{0.8}\selectfont]
...
\end{tcolorbox}

\begin{tcolorbox}[breakable, title=Step 4 ,colback=gray!5,colframe=nmgray!75!black,before skip=1pt, after skip=1pt,fontupper=\linespread{0.8}\selectfont]
{\footnotesize
\textbf{Paper List:}

[\textcolor{orange}{2112.09106}] (\textcolor{blue}{1.00}) [\textcolor{green}{EXP}] RegionCLIP: Region-based Language-Image Pretraining 

Abstract: CLIP is ineffective for region-level object detection due to the domain shift between whole-image–text alignment and region–text alignment.
RegionCLIP extends CLIP to learn region-level visual representations, significantly improving open-vocabulary and zero-shot object detection performance.
...

[\textcolor{orange}{2104.13921}] (\textcolor{blue}{0.99}) [\textcolor{green}{EXP}] Open-vocabulary Object Detection via Vision and Language Knowledge Distillation 

Abstract: We propose ViLD, a vision-language knowledge distillation framework that transfers open-vocabulary knowledge from a pretrained image-text model to a two-stage detector.
ViLD achieves strong generalization on multiple detection benchmarks and outperforms prior methods on novel categories without additional finetuning.
...

[\textcolor{orange}{1908.02265}] (\textcolor{blue}{0.54}) [\textcolor{red}{NEW}] ViLBERT: Pretraining Task-Agnostic Visiolinguistic Representations for Vision-and-Language Tasks 

Abstract: We propose ViLBERT, a task-agnostic vision-and-language pretraining model with a two-stream architecture and co-attentional transformers.
Pretrained on large-scale image-text data, ViLBERT achieves state-of-the-art performance across multiple vision-language tasks with minimal task-specific modification.
...

(...More Papers...)

[\textcolor{orange}{2301.12597}] (\textcolor{blue}{0.05}) [\textcolor{red}{NEW}] BLIP-2: Bootstrapping Language-Image Pre-training with Frozen Image Encoders and Large Language Models 

Abstract: We propose BLIP-2, an efficient vision-language pretraining framework that leverages frozen image encoders and frozen large language models.
With a lightweight querying transformer, BLIP-2 achieves state-of-the-art performance and strong zero-shot capabilities with significantly fewer trainable parameters.
...

\textbf{Analysis:}

The user cares about the studies that leverage CLIP's capabilities, and we have already identified the primary paper on RegionCLIP (arXiv:2112.09106), which is now marked as [EXP] (expanded). This paper is the central focus and has been used as a seed for expansion via the expand tool (e.g., expanded papers like 2104.13921, 2308.02299, etc.).

Currently, the paper list includes several relevant papers on vision-language pre-training and region-based models, such as:
RegionBLIP (2308.02299): focuses on regional comprehension in multi-modal models.
UrbanCLIP (2310.18340): uses text-enhanced urban region profiling with CLIP-like training.
Oscar (2004.06165): object-semantic alignment pre-training.
FILIP (2111.07783): fine-grained language-image pre-training with token-wise alignment.
STAIR (2301.13081): sparse text and image representation.

These papers are closely related in theme, focusing on regional or fine-grained visual understanding in vision-language settings, which supports the broader context of RegionCLIP as a foundation. However, many of these have been already expanded (marked [EXP]), so further expansion may not yield new relevant papers.

To deepen the exploration, we should:
Identify missing papers from the primary RegionCLIP literature by searching on related keywords to complement what we already found.
Expand on other key papers that are relevant but not yet expanded, such as:
STAIR (2301.13081): sparse representations, strong performance in retrieval.
FILIP (2111.07783): fine-grained alignment, strong localization.
BLIP-2 (2301.12597): foundational MLLM, influential in regional vision-language tasks.

Avoid expanding on already expanded papers (e.g., 2112.09106, 2104.13921, 2308.02299, 2310.18340, 2305.05095) to avoid redundancy.

Therefore, the next actionable steps should include:
Searching for papers related to `fine-grained region understanding in vision-language models'' to complement existing ones. 
Expanding on key under-explored candidates like FILIP (2111.07783) and BLIP-2 (2301.12597), as they offer strong theoretical or practical advancements in regional language-image alignment.

This will ensure we cover both complementary and emerging directions in the region-based vision-language pre-training space.

\textbf{Tool Call:}
\begin{PromptCode}
{"name": "search", "arguments": {"query": "fine-grained region understanding in 
vision-language models"}}
{"name": "expand", "arguments": {"arxiv_id": "2111.07783"}}
{"name": "expand", "arguments": {"arxiv_id": "2301.12597"}}
\end{PromptCode}
}
\end{tcolorbox}

\begin{tcolorbox}[breakable, title=More Steps ,colback=gray!5,colframe=nmgray!75!black,before skip=1pt, after skip=1pt,fontupper=\linespread{0.8}\selectfont]
...
\end{tcolorbox}

\begin{tcolorbox}[breakable, title=Final Result ,colback=gray!5,colframe=nmgray!75!black,before skip=1pt, after skip=1pt,fontupper=\linespread{0.8}\selectfont]
{\footnotesize

\textbf{Paper ID:} 2112.09106
\begin{itemize}
    \item \textbf{Title:} RegionCLIP: Region-based Language-Image Pretraining
    \item \textbf{Abstract:} ...
    \item \textbf{Score:} 0.9957
\end{itemize}

\textbf{Paper ID:} 2104.13921
\begin{itemize}
    \item \textbf{Title:} Open-vocabulary Object Detection via Vision and Language Knowledge Distillatio
    \item \textbf{Abstract:} ...
    \item \textbf{Score:} 0.9949
\end{itemize}

(...More Papers...)

\textbf{Paper ID:} 2203.16513
\begin{itemize}
    \item \textbf{Title:} PromptDet: Towards Open-vocabulary Detection using Uncurated Images
    \item \textbf{Abstract:} ...
    \item \textbf{Score:} 0.5027
\end{itemize}

}
\end{tcolorbox}

% \section{Example Appendix}
% \label{sec:appendix}

% This is an appendix.

\end{document}